\documentclass{article} %
\usepackage{colm2024_conference}

\usepackage{pdfpages}
\usepackage{microtype}
\usepackage{hyperref}
\usepackage{url}
\usepackage{booktabs}
\usepackage{paralist}
\definecolor{darkblue}{rgb}{0, 0, 0.5}
\hypersetup{colorlinks=true, citecolor=darkblue, linkcolor=darkblue, urlcolor=darkblue}

\usepackage{bbm}
\usepackage{listings}
\usepackage{graphicx}
\usepackage{caption}
\usepackage{hyperref}
\usepackage{subcaption}
\usepackage{amsmath}
\usepackage{booktabs}
\usepackage{multirow}
\usepackage{graphicx}
\usepackage{epsfig}
\usepackage{amsmath}
\usepackage{amssymb}
\usepackage{enumitem}
\usepackage{placeins}
\usepackage{multirow}
\usepackage{booktabs}
\usepackage{rotating}
\usepackage{cancel}
\usepackage{setspace}
\usepackage{eso-pic}

\def\x{\mathbf{x}}

\def\y{\mathbf{y}}

\DeclareMathOperator*{\argmax}{arg\,max}

\definecolor{trolleygrey}{rgb}{0.5, 0.5, 0.5}
\definecolor{darkpastelgreen}{rgb}{0.01, 0.75, 0.24}
\definecolor{darkpink}{rgb}{0.91, 0.33, 0.5}
\definecolor{alizarin}{rgb}{0.82, 0.1, 0.26}
\definecolor{americanrose}{rgb}{1.0, 0.01, 0.24}

\colmfinalcopy

\title{AI-generated text boundary detection with RoFT}

\author{
 Laida Kushnareva$^{1}$, Tatiana Gaintseva$^{2}$, \\ German Magai$^{3,4}$, 
 Serguei Barannikov$^{5,6}$, Dmitry Abulkhanov, Kristian Kuznetsov$^{5}$, \\ Eduard Tulchinskii$^{5}$,
 Irina Piontkovskaya$^{1}$,  Sergey Nikolenko$^{7}$
 \\
  \\ \textsuperscript{1}AI Foundation and Algorithm Lab, Russia;  
  \\ \textsuperscript{2} Digital Environment Research Institute, Queen Mary University of London, UK;
  \\ \textsuperscript{3} HSE University, Russia;  \textsuperscript{4} Noeon Research, Japan; \\ \textsuperscript{5} Skolkovo Institute of Science and Technology, Russia;
  \\ \textsuperscript{6} CNRS, Université Paris Cité, France;
\\\textsuperscript{7} St. Petersburg Department of the Steklov Institute of Mathematics, Russia \\
\\
\small{
    \textbf{Correspondence:} \href{mailto:kushnareva.laida@gmail.com}{kushnareva.laida@gmail.com}}
}

\begin{document}
\maketitle
\begin{abstract}
Due to the rapid development of large language models, people increasingly often encounter texts that may start as written by a human but continue as machine-generated. Detecting the boundary between human-written and machine-generated parts of such texts is a challenging problem that has not received much attention in literature. We attempt to bridge this gap and examine several ways to adapt state of the art artificial text detection classifiers to the boundary detection setting. We push all detectors to their limits, using the Real or Fake text benchmark that contains short texts on several topics and includes generations of various language models. We use this diversity to deeply examine the robustness of all detectors in cross-domain and cross-model settings to provide baselines and insights for future research. In particular, we find that perplexity-based approaches to boundary detection tend to be more robust to peculiarities of domain-specific data than supervised fine-tuning of the RoBERTa model; we also find which features of the text confuse boundary detection algorithms and negatively influence their performance in cross-domain settings.

\end{abstract}

\section{Introduction}

Artificial text detection (ATD) is a very difficult problem in real life, where machine-generated text may be intertwined with human-written text, lightly edited, or pad out human-generated prompts. However, in literature ATD is usually formulated in a simpler way, with 
a dataset of text samples labeled as either entirely human-written or entirely machine-written, so the detection problem can be safely treated as binary classification. Moreover, the models for this binary classification are often developed and trained to detect a particular type of generator, e.g., text produced by a specific large language model (LLM) (\citet{Adaku_2023}). This is in stark contrast with how we may encounter artificially created text in real life, where documents partially written by humans and partially generated by LLMs already abound. This setting is much more complex and much less researched.

\begin{figure*}[!t]\centering
\includegraphics[width=\textwidth]{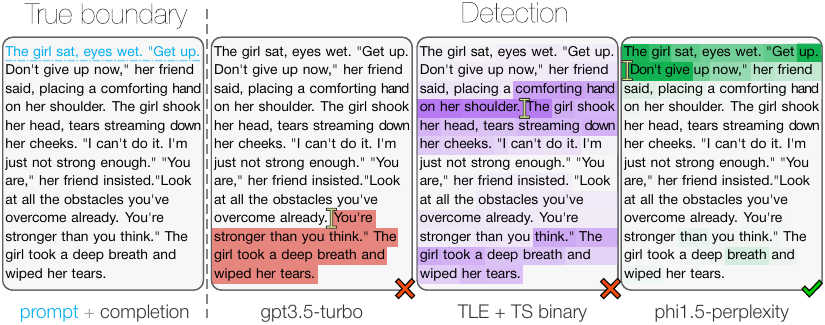}
\caption{Sample input from the ROFT-chatgpt dataset. The rightmost part of the picture shows the true bondary between human-written and machine-generated text. Other three parts are colored according to the features, extracted by various classifiers. Here, gpt3.5 predicts that only two last sentences are generated; TLE + TS binary predicts that generation starts from the second sentence; phi1.5 perplexity predicts that generation starts from the third sentence which is correct answer. See Section ~\ref{sec:method} for detailed explanation of all the methods and Appendix~\ref{sec:misclassified_examples} for more examples from the dataset.}
\label{fig:overview}
\end{figure*}

In this work, we experiment with a lesser known dataset called RoFT (Real Or Fake Text), collected by~\citet{10.1609/aaai.v37i11.26501}. Each text in this dataset consists of ten sentences, where the first several sentences are human-written and the rest are machine-generated starting from this prompt, mainly by models from the GPT family~\citep{radford2019language,NEURIPS2020_1457c0d6}. We consider several techniques developed for binary ATD, modifying them for this more complex boundary detection setting; e.g., following \citet{tulchinskii2023intrinsic} we adapt intrinsic dimension estimation which is currently considered to be the most robust method for cross-domain and cross-model binary detection. Fig.~\ref{fig:overview} shows a sample input from RoFT and results of the considered models.
Our primary contributions are as follows:
\begin{inparaenum}[(1)]
    \item we evaluate five approaches to adapt perplexity-based detectors to the task of detecting the boundary between human-written and machine-generated text, discussing the differences in their behavior compared to binary ATD and providing a comprehensive analysis of how perplexity scores react to the machine--human transition in the text; we also compare several backbones for the perplexity evaluation to show which model works the best for this purpose;
    \item we present evidence that perplexity scores obtained from small language model, trained on data generated by bigger LLM, provide strong features for boundary detection models, leading to relatively high accuracy of detection and cross-domain robustness related to other detection methods;   
    \item we introduce and evaluate two ways to adapt the classifiers based on intrinsic dimension estimation for this task, showing how the time series analysis can be used for extracting useful information from Transformer representations; 
    \item we show how the robustness of boundary detectors (including the fully-tuned RoBERTa baseline) to domain shift depends on the particular properties of the domains; we study the properties of the dataset itself and their effect on the performance of every approach; %
    \item we enrich the RoFT dataset with \emph{GPT-3.5-turbo}\footnote{\url{https://platform.openai.com/docs/model-index-for-researchers}} (ChatGPT) generation samples; we share this new dataset with the community, establish baselines, and analyze the behavior of our detectors on it.\footnote{Our official repository: \url{https://github.com/SilverSolver/ai_boundary_detection.git}}
\end{inparaenum}

We hope that this work will encourage further research both in boundary detection for texts that are partially human-written and partially generated and analyzing how inner representations of Transformer-based models react to such transitions; the latter direction may also help interpretability research.
The rest of the paper is organized as follows. In Section~\ref{sec:related} we survey related work, and Section~\ref{sec:method} introduces the methods we have applied for artificial text boundary detection. Section~\ref{sec:eval} presents a comprehensive evaluation study on the RoFT and RoFT-chatgpt datasets. Section~\ref{sec:analysis} presents a detailed discussion and analysis of our experimental results, and Section~\ref{sec:concl} concludes the paper.

\section{Related Work}\label{sec:related}

\textbf{Artificial text detection} (ATD) is a well-studied task with plenty of already considered approaches; see, e.g., the surveys by \citet{yang2023survey,Adaku_2023,wu2023survey}. In our work, we mostly concentrate on adapting two groups of methods to artificial text boundary detection: methods based on perplexity estimation since they are the most widely used and methods based on topological data analysis (TDA) since they have shown promising robustness to domain shift and model shift (see below).%
\emph{Perplexity-based methods}
are based on the idea that human-written text tends to be more unpredictable and varied compared to AI-generated text which may follow more predictable patterns, i.e., has a lower perplexity score; among them we note GPTZero \citep{weber2023testing}, Sniffer \citep{li2023origin}, SeqXGPT \citep{wang-etal-2023-seqxgpt}, and LLMDet \citep{wu2023llmdet}.
\emph{Topological data analysis} (TDA) for ATD
is inspired by the results of \citet{kushnareva-etal-2021-artificial} and \citet{tulchinskii2023intrinsic}. The latter has shown that for many artificial text generators
the intrinsic dimensions (PHD) of RoBERTa embeddings of the texts created by these generators are typically smaller than those of the texts written by humans; %
also TDA has proven to be useful for closely related tasks of fake news detection \citep{osti_10422762}, authorship attribution \citep{elyasi2019introduction}, and detection of synthetic speech \citep{tulchinskii23_interspeech}.

\textbf{Style change detection and authorship attribution}.
ATD can be considered as a special case of authorship attribution (AA), where the LLM is one author and the human is another, and the task is to determine the authorship of different parts of the text; it is also similar to the style change detection problem in multi-author documents \citep{Zangerle2021OverviewOT}. \citet{jones2022you} show that LLM models can successfully imitate human style and deceive existing popular online AA methods. However, \citet{venkatraman2023gpt} propose an approach based on the principle of uniform information density that can detect the authorship of LLM.
State of the art style change detectors \citep{DBLP:conf/clef/LinCTL22,Jiang2022StyleCD,DBLP:conf/clef/LaoMYYYTL22,Iyer2020StyleCD} are based on Transformer-based encoders such as BERT \cite{devlin-etal-2019-bert}, RoBERTa \cite{liu2019roberta}, AlBERT \cite{DBLP:conf/iclr/LanCGGSS20}, and ELECTRA \cite{clark2020electra}. Therefore, we use RoBERTa model as a baseline and a source of embeddings.

\textbf{RoFT (Real Or Fake Text)}. %
Our main dataset \citep{dugan-etal-2020-roft,10.1609/aaai.v37i11.26501}
originates from a website called RoFT\footnote{\url{http://www.roft.io/}} developed as a tool to analyze how \emph{humans} detect generated text, including an extended study of whether and how they can explain their choice, when they say that a text sample is machine-generated, and how humans can learn to recognize machine-generated text better. Every user of this website can choose a topic (``Short Stories'', ``Recipes'', ``New York Times'', or ``Presidential Speeches'') and start the following ``game''.
The player sees ten sentences one by one. The first sentence is always written by a human, but for each subsequent sentence the player must determine whether it is machine-generated or human-written. If the player believes the text was AI-generated, they should explain why they think so. If they guess machine generation before the true boundary, they earn zero points. Otherwise, they earn $5 - x$ points, where $x$ is the number of sentences after the correct boundary. This is actually a harder problem than just boundary detection since the player does not see the full text, and the scoring function is skewed.
As a result,
\citet{10.1609/aaai.v37i11.26501} created a dataset also known as RoFT, where
every sample consists of ten sentences, starting with a human-written prompt and continuing by some language model; the data shows the true boundary, human prediction of it, the generator model, an explanation provided by the player, and information about the player.
The original RoFT contains generations from GPT-2~\citep{radford2019language}, GPT-2 XL, GPT-2 finetuned on the ``Recipes'' domain, GPT-3.5 (davinci), CTRL~\citep{keskarCTRL2019} with control code ``nocode'' and with control code ``Politics'', and baseline, where instead of an LLM-generated continuation the passage transitions to a completely different news article selected at random.
RoFT has been mostly used to investigate how humans detect artificially generated texts manually \citep{clark-etal-2021-thats};
\citet{cutler2021automatic} provided the first baselines on automatically solving RoFT, comparing several shallow classification and regression models based on RoBERTa and SRoBERTa~\citep{reimers-gurevych-2019-sentence} embeddings collected from the last layer of these models. 
However, the cross-domain and cross-model settings in their research were very limited. In fact, \citet{cutler2021automatic} call ``Out of Domain'' (OOD) classifiers trained on \emph{all} available data and then evaluated on a given subset, and they call ``In Domain'' (ID) classifiers that had been trained and evaluated on the same subset. Besides, they did not analyze all generators and domains within this cross-domain setting. In this work, we concentrate on cross-domain and cross-model settings and interpretability, evaluating boundary detectors on unseen generators (models) and topics (domains). 

\textbf{Other related work}.
\citet{zeng2023towards} used a TriBERT-based approach for artificial text boundary detection in student essays for educational purposes, but with no cross-domain problem setting.
\cite{wang-etal-2023-seqxgpt} address mixed human and AI generations by sentence-level classification, but they do it with a \textit{white-box detector} that needs access to the generator model, which severely limits applicability.
Recently, a SemEval workshop\footnote{\url{https://aclanthology.org/venues/semeval/}} introduced a small artificial text boundary detection dataset, namely subtask~C of Task~8\footnote{\url{https://github.com/mbzuai-nlp/SemEval2024-task8}}. The subtask asks to predict the word where human text ends and artificial text begins. However, to this moment there is no information on (a) which model was used to generate the samples, (b) what are the particular domains or topics of the generations, and (c) any human baselines. Existing technical reports on the competition \citep{spiegel2024kinit,datta2024semeval2024,rad2024rfbes} concentrate on other subtasks of Task~8 such as binary ATD and LLM-authorship attribution, but not the boundary detection subtask.

\section{Approach}\label{sec:method}

We consider several different approaches, including a 
multilabel classification framework, as proposed by \citet{cutler2021automatic}, where the label of a text corresponds to the number of the first generated sentence,
time series analysis that slides a window over the text tokens,
and regression methods that minimize the MSE between true and predicted boundaries.
We design our classifiers based on features that have been successfully used in prior works on ATD~\citep{solaiman2019release, mitchell2023detectgpt}. We also introduce a new baseline based on sentence lengths.
Below, we consider all of these approaches in detail.

\textbf{RoBERTa classifier}.
Unlike \citet{cutler2021automatic} who process each sentence separately, we fine-tune the RoBERTa model \citep{liu2019roberta} to represent the entire text sample via the [CLS] vector. This is the only case in our work where we apply full model fine-tuning. For other methods, we use simpler classifiers such as logistic regression (LR) or gradient boosting (GB) \citep{4a848dd1-54e3-3c3c-83c3-04977ded2e71} trained on various features extracted directly from larger models (e.g., Transformer-based LLMs), with no updates to the larger model's weights.

\textbf{Perplexity from generative LLMs}. 
In our perplexity experiments, we implement the \textit{black-box approach}, which means that a single model is used for all data to compute sentence-wise perplexity. This is more practical than the method of \citet{cutler2021automatic} who used perplexity scores from the original generator model, which 
may be infeasible if the generator is unknown, in particular in cross-model scenarios. Even for known generators, the exact \textit{version} of the model is usually unknown if they are available via API, which harms the performance of \textit{white-box} detectors.
To derive perplexity features, we calculate the log likelihood of each token via LLM, and derive sentence-level \textit{mean} and \textit{std} values. As the perplexity estimator, we use several models such as GPT-2~\citep{radford2019language}, Phi-1 ~\citep{gunasekar2023textbooks}, Phi-1.5 ~\citep{li2023textbooks}, Phi-2~\citep{li2023phi2}, and LLaMA-2-7B~\citep{touvron2023llama}. Our underlying hypothesis here is that texts generated by models of similar architecture such as GPT-2 or GPT-3.5 might appear more ``natural'' to these models, as reflected by their likelihood scores. Our findings corroborate this assumption (see Appendix~\ref{sec:approaches_appendix}).
On top of sentence-wise perplexity features, we train either a \emph{classifier} (logistic regression and gradient boosting) or a \emph{regressor} that predicts boundary values (GB regressor); the regression setting takes advantage of the sequential nature of the task, minimizing the discrepancy in labels.

\textbf{DetectGPT}.
The DetectGPT framework \citep{mitchell2023detectgpt} proposes a more nuanced perplexity-based scoring function. It involves perturbing a text passage and comparing log-probabilities between original and altered texts. This score then serves as an input to a classification model, with GPT-2 as the base model and T5-Large \citep{JMLR:v21:20-074} generating the perturbations.

\textbf{Length-Based Baseline}.
Since we have observed a statistical difference in sentence length distributions between human-written and generated texts (see Fig.~\ref{fig:roft-by-model}), we leverage sentence lengths as a simple baseline feature. This baseline allows us to gauge the effectiveness of a classifier in identifying boundaries without semantic understanding.

\textbf{Topological Time Series (TTS)}.
Inspired by \citet{tulchinskii2023intrinsic}, we explore the potential of topological features based on intrinsic dimensionality (ID); we provide an introduction to topological data analysis (TDA), including the definitions of features, in Appendix~\ref{sec:id_appendix}. 
We hypothesize that geometric variations in token sequences can help identify AI-generated text, so we introduce models that process TDA-based features treating them as time series.
For every text, we slide a window of $H = 100$ tokens (step size $S = 1$) over RoBERTa token embeddings and find the intrinsic dimension (PHD) of the points within the window, as shown in Fig.~\ref{fig:three graphs} \citep{schweinhart2020fractal}. Window size $100$ was chosen as a minimal size where PHD estimator stayed stable. The time series are then classified with a multi-label SVM with the global alignment kernel (GAK) \citep{cuturi2011fast}, where label correspond to a number of the first generated sentence; this method is called ``PHD + TS ML'' in the tables.

\textbf{Topological binary classification}.
In this approach, we train a binary classifier to distinguish between \emph{fake} and \emph{natural} text atop a specific predictor. For the base predictor, we employ intrinsic dimensionality calculated over a sliding window of $H = 20$ tokens (step size $S = 5$)\footnote{Stride size 5 was chosen empirically after the preliminary experiments. In these preliminary experiments we found out that for smaller stride sizes the series became too long for the Time Series SVM algorithm, and it wasn't able to converge.}. The TLE (tight local) intrinsic dimension estimator was chosen due to its robust performance on small data samples; window size $20$ was taken directly from \citet{amsaleg:hal-02125331}, as the authors claim that their algorithm is stable enough “for `tight' localities consisting of as few as 20 sample points”. For the base classifier, we use gradient boosting trees \citep{friedman2000greedy, FRIEDMAN2002367}.
To translate probabilities predicted with a binary classifier into a specific boundary that separates real and fake text, we determine the final label by maximizing $\y \mapsto \argmax_{I \in \mathcal{I}} s_{I}(\y, \x)$, where the score function $s_{I}$ is defined as $s_I(\y, \x) = - \sum_{j}^n I^{j}_\y \log \hat{p}(\x^j)$, and $\hat{p}(\x)$ are predictions from the base binary classifier. In this context, we use a binary indicator set $I \in \mathbf{1}^{n}$ consisting of 
a chunk of $k$ zeros followed by another chunk of $n-k$ ones. This method is called ``TLE + TS Binary'' in the tables.

\textbf{Zero-shot detection}. As an additional baseline, we tried zero-shot boundary detection with two strongest black-box models: gpt-3.5-turbo and gpt-4. We describe the problem to them with a natural language prompt which is provided in Appendix \ref{sec:chatgpt}.

\section{Experimental evaluation}\label{sec:eval}

\textbf{Dataset preparation and analysis}.
In all experiments, the task is to detect the exact boundary where a text passage that started as human-written transits to machine generation.
In addition to RoFT \citep{dugan-etal-2020-roft}, we created a new dataset called \textbf{RoFT-chatgpt},
where the same human prompts
are continued with the \emph{gpt-3.5-turbo} model.
RoFT-chatgpt is designed to be more challenging for artificial text boundary detection while preserving the basic statistical properties of RoFT such as the label distribution.
We removed duplicates that were different only in human-predicted labels; 
for \emph{RoFT-chatgpt}, we also removed samples containing ``As an AI language model...'' and short failed generations. As a result, we retained $8943$ samples from the original RoFT and $6940$ samples in \emph{RoFT-chatgpt}.
Note that distributions of sentence lengths vary significantly across subdomains, as shown in Fig.~\ref{fig:roft-by-model} and Figs.~\ref{fig:roft-by-topic},~\ref{fig:roft-chatgpt-by-topic} in Appendix~\ref{sec:dataset_appendix}
(see also a discussion in Section~\ref{sec:analysis}).

\begin{table*}[!t]
  \centering\setlength{\tabcolsep}{4pt}
  \caption{Boundary detection results. \textbf{Bold} shows the best method, \underline{{underlined}} - second best, \textit{italic} for values outperforming the human baseline.}\label{tab:results}

  \resizebox{0.92\linewidth}{!}{
  \begin{tabular}{l|lll|lll}\toprule
    \multicolumn{1}{c|}{\textbf{Method}} & \multicolumn{3}{c|}{\textbf{RoFT}} & \multicolumn{3}{c}{\textbf{RoFT-chatgpt}} \\
    \multicolumn{1}{c|}{} & Acc & SoftAcc1 & MSE & Acc & SoftAcc1 & MSE \\
    \midrule
    \multicolumn{1}{l|}{RoBERTa + SEP} & \textbf{49.64} \% & \textbf{79.71} \% & \textbf{02.63} & \textbf{54.61} \% & \textbf{79.03} \% & \textbf{03.06} \\
    \multicolumn{1}{l|}{RoBERTa} &  \underline{{46.47}} \% &  \underline{{74.86}} \% &  \underline{{03.00}} &  39.01 \% &  \underline{{75.18}} \% &  03.15 \\
    \midrule
    \multicolumn{1}{l|}{Based on Perplexity} & & & &  & & \\
    \hspace{20pt}\emph{DetectGPT} + GB classifier & 19.79 \% & 37.40 \% & \textit{08.35} & 21.69 \% & 43.52 \% & 06.87 \\
    \hspace{20pt}\emph{DetectGPT} + LR classifier &  19.45 \% & 33.82 \% & \textit{09.03} & 15.35 \% & 41.43 \% & 07.22 \\
    \hspace{20pt}\emph{GPT-2 Perpl.} + GB classifier & \textit{24.25} \% & \textit{47.23} \% & \textit{11.68} & 34.94 \% & 59.80 \% & 07.46 \\
    \hspace{20pt}\emph{GPT-2 Perpl.} + GB regressor & 12.58 \% & 36.67 \% & \textit{06.89} & 19.74 \% & 54.03 \% & 04.89 \\
    \hspace{20pt}\emph{GPT-2 Perpl.} + LR classifier & \textit{23.75} \% & \textit{42.15} \% & 15.80 & 33.50 \% & 57.56 \% & 09.25\\
     \hspace{20pt}\emph{Phi-1 Perpl.} + GB classifier & 20.9 \% & 42.7 \% & 12.92 & 36.5 \% & 56.0 \% & 09.13 \\
     \hspace{20pt}\emph{Phi-1 Perpl.} + GB regressor & 10.7 \% & 33.0 \% & 07.51 & 17.4 \% & 49.9 \% & 05.51 \\
     \hspace{20pt}\emph{Phi-1 Perpl.} + LR classifier & 19.2 \% & 38.3 \% & 16.26 & 31.0 \% & 53.9 \% & 09.76 \\
     \hspace{20pt}\emph{Phi-1.5 Perpl.} + GB classifier & \textit{29.7} \% & \textit{55.0} \% & \textit{09.50} & \underline{{50.8}} \% & 74.1 \% & 04.59 \\
     \hspace{20pt}\emph{Phi-1.5 Perpl.} + GB regressor & 17.1 \% & \textit{44.6} \% & \textit{06.11} & 32.0 \% & 71.0 \% & \underline{{03.07}} \\
     \hspace{20pt}\emph{Phi-1.5 Perpl.} + LR classifier & \textit{27.0} \% & \textit{49.5} \% & \textit{11.90} & 47.3 \% & 72.7 \% & 04.77 \\
     \hspace{20pt}\emph{Phi-2 Perpl.} + GB classifier & \textit{35.4} \% & \textit{58.3} \% & \textit{08.80} & 48.4 \% & 70.3 \% & 05.39 \\
     \hspace{20pt}\emph{Phi-2 Perpl.} + GB regressor & 15.6 \% & \textit{43.7} \% & \textit{05.83} & 30.8 \% & 68.7 \% & 03.33 \\
     \hspace{20pt}\emph{Phi-2 Perpl.} + LR classifier & \textit{25.8} \% & \textit{49.5} \% & \textit{11.78} & 44.4 \% & 71.5 \% & 05.42 \\
     \hspace{20pt}\emph{LLaMA-2-7B Perpl.} + GB classifier & 22.3 \% & \textit{42.0} \% & 14.79 & 41.5 \% & 65.0 \% & 06.32 \\
     \hspace{20pt}\emph{LLaMA-2-7B Perpl.} + GB regressor & 11.4 \% & 32.9 \% & \textit{7.98} & 23.6 \% & 60.3 \% & 03.89 \\
     \hspace{20pt}\emph{LLaMA-2-7B Perpl.} + LR classifier & 19.4 \% & 34.7 \% & 19.75 & 29.7 \% & 58.4 \% & 07.90 \\
     \midrule
     \multicolumn{1}{l|}{Based on TDA} & & & &  & & \\
     \hspace{20pt}PHD + TS ML & \textit{23.50} \% & \textit{46.32} \% & 14.14 & 17.29 \% & 35.81 \% & 14.45 \\
     \hspace{20pt}TLE + TS Binary & 12.58 \% & 30.41 \% & 22.23 & 20.02 \% & 34.58 \% & 18.52  \\
    \midrule
     \multicolumn{1}{l|}{Zero-shot} & & & &  & & \\
     \hspace{20pt}gpt-3.5-turbo & 02.5 \% & 17.3 \% & 30.3 & 07.1 \% & 23.9 \% & 25.5 \\
     \hspace{20pt}gpt-4 & 06.4 \% & 22.0 \% & 27.0 & 09.1 \% & 25.0 \% & 24.1 \\
    \midrule
    \multicolumn{1}{l|}{\emph{Length} + GB} & 14.64 \% & 33.43 \% & 16.55  & 25.72 \% & 46.18 \% & 18.99 \\
    \multicolumn{1}{l|}{Majority class} & 15.26 \% & 25.43 \% & 27.58 & 13.83 \% & 24.42 \% & 26.46 \\
    \multicolumn{1}{l|}{SRoBERTa \citep{cutler2021automatic}} & 42 \% & --- & --- & \multicolumn{3}{c}{---} \\
    \multicolumn{1}{l|}{Human baseline} & 22.62 \% & 40.31 \% & 13.88 & \multicolumn{3}{c}{---}\\\bottomrule
\end{tabular}
}
\end{table*}

\textbf{Artificial text boundary detection}.
Table~\ref{tab:results} presents the main results of our experiments 
on the RoFT and RoFT-chatgpt datasets; the
majority class prediction is
the last ($9$th) class of fully human-written texts. On the original RoFT, we
also include 
the human baseline and the best result reported by \citet{cutler2021automatic}, obtained by a classifier built upon concatenated SRoBERTa embeddings of each sentence.
We report the accuracy (\emph{Acc}) and two metrics that account for the sequential nature of boundary detection: 
soft accuracy, i.e., the percentage of predictions that differ from the correct label by at most one (\emph{SoftAcc1}; indeed, Fig.~\ref{fig:heatmaps} in Appendix~\ref{sec:approaches_appendix} shows that many misclassifications point to an adjacent class, which may be acceptable in real world applications), and mean squared error (MSE). More precisely, $SoftAcc1 = \frac{1}{n}\sum^{n}_{i = 1}{\mathbbm{1}_{\{|y_i - \hat{y_i}| \le 1\}}}$ and $MSE = \frac{1}{n}\sum^{n}_{i = 1}(y_i - \hat{y}_i)^2$, where $y_i$ - predictions of the algorithm, $\hat{y_i}$ - true labels, $n$ - amount of the examples.

First, on the original \emph{RoFT} RoBERTa-based classifiers outperform others 
by a large margin (14\% accuracy)
and also significantly outperform the previously best reported SRoBERTa \citep{cutler2021automatic}. This model also provides the lowest MSE ($0.03$) among all methods. We note, however, that our RoBERTa classifier has significantly more trainable parameters than any other method in the table because no other approaches require LM fine-tuning.
Second, topological and perplexity features improve over the human baseline. Perplexity-based classifiers are the best in terms of accuracy, while the perplexity regressor provides good MSE values (recall from Section~\ref{sec:related}, however, that humans were solving a harder problem with a somewhat different objective).
Third, RoBERTa's accuracy on \emph{RoFT-chatgpt} drops by 6\% compared to RoFT, while soft accuracy and MSE are roughly the same, but 
the opposite holds for perplexity-based methods: on \emph{RoFT-chatgpt} their results are much \emph{better}.
The reason for this might be that we used GPT-like models for perplexity calculation and GPT-3.5-turbo to generate fake samples in \emph{RoFT-chatgpt}. 
However, we used smaller models to detect text generated by a larger model (GPT-3.5) and got second-best results among other approaches, despite \citet{mitchell2023detectgpt} reporting that smaller models are not capable of detecting text generated by larger models. 
Moreover, when analyzing the performance of different perplexity estimators, we do not observe a clear correlation of the performance with the size of the estimator. The largest LLaMA model with 7B parameters works much worse than Phi-1.5 (1.3B) and Phi-2 (2.7B) on both datasets and slightly worse than GPT-2 (1.5B) on \emph{RoFT}. For \emph{RoFT-chatgpt}, the best perplexity estimator is a relatively small Phi-1.5 model, outperforming a GPT-2 estimator of comparable size by 15\%. This impressive difference might be related to training data: Phi-1.5 was trained on GPT-3.5 generations. This suggests that small language models trained on data generated by LLMs, may appear a good perplexity-based detector of text generated by those particular LLMs. 
The baseline length-based classifier also improves its accuracy significantly (by 1.8x) when transferring to \emph{RoFT-chatgpt}. We hypothesize that this kind of shallow feature emerges in ChatGPT generation and makes the task easier (see also Section~\ref{sec:analysis}). 
The other perplexity-based approach, DetectGPT, shows lower accuracy 
on both datasets. \citet{mitchell2023detectgpt} note that DetectGPT
can detect whether the sample is generated by a specific base model, but we use several models in our setup, and text samples may be too short for this approach. On the other hand, DetectGPT has quite good MSE values, close to the regression approach that optimizes MSE directly.

\begin{table*}[!t]
\centering
\setlength{\tabcolsep}{0.15em}
  \caption{
  Accuracy for leave-one-out cross-domain evaluation on \emph{RoFT-chatgpt}. $\vartriangle$ and $\triangledown$ show relative change from the model's \textit{in-domain} score to the human score; $\blacktriangle$ and $\blacktriangledown$, relative change from the \textit{out-of-domain} score to the \textit{in-domain} score. $\textcolor{darkpastelgreen}{Green}$ highlights improvements, $\textcolor{darkpink}{red}$, deteriorations. The table shows only perplexity methods based on perplexity of Phi1.5 and Phi2; for other perplexity backbones, see Appendix~\ref{sec:appendix_crossdomain}.
  }
  \label{tab:cross_roft}

\resizebox{\textwidth}{!}{
\begin{tabular}{@{}lll|ll|ll|ll|ll|r@{}}
\toprule
&
 \multicolumn{2}{c|}{} &
  \multicolumn{2}{c|}{\textbf{Pres. Speeches}} &
  \multicolumn{2}{c|}{\textbf{Recipes}} &
  \multicolumn{2}{c|}{\textbf{New York Times}} &
  \multicolumn{2}{c|}{\textbf{Short Stories}} &
  \small \textbf{Avg} \\ %
\textbf{Pred.} &
  \textbf{Model} &
  \textbf{Context} &
  IN $\uparrow$ &
  OUT $\uparrow$ &
  IN $\uparrow$ &
  OUT $\uparrow$ &
  IN $\uparrow$ &
  OUT $\uparrow$ &
  IN $\uparrow$ &
  OUT $\uparrow$ &
  \multicolumn{1}{c}{$\Delta \downarrow$} \\ 
  \midrule
  Text &
    RoBERTa SEP&
  global &
  $\underline{57.3} \textcolor{darkpastelgreen}{\scriptstyle \vartriangle 153 \%} $ &
  $31.4 \textcolor{darkpink}{\scriptstyle \blacktriangledown 45 \%} $ &
  $43.2 \textcolor{darkpastelgreen}{\scriptstyle \vartriangle 91 \%} $ &
  $13.1 \textcolor{darkpink}{\scriptstyle \blacktriangledown 70 \%} $ &
  $\underline{53.2} \textcolor{darkpastelgreen}{\scriptstyle \vartriangle 135 \%} $ &
  $38.1 \textcolor{darkpink}{\scriptstyle \blacktriangledown 28 \%} $ &
  $\underline{54.3} \textcolor{darkpastelgreen}{\scriptstyle \vartriangle 140 \%} $ &
  $28.6 \textcolor{darkpink}{\scriptstyle \blacktriangledown 
47 \%} $ &
  $-48\%$ \\
Text &
  RoBERTa &
  global &
  $\textbf{67.6}   \textcolor{darkpastelgreen}{\scriptstyle \vartriangle 199\%} $ &
  $36.3   \textcolor{darkpink}{\scriptstyle \blacktriangledown 46\%} $ &
  $53.7   \textcolor{darkpastelgreen}{\scriptstyle \vartriangle 134\%} $ &
  $14.8   \textcolor{darkpink}{\scriptstyle \blacktriangledown 72\%} $ &
  $\textbf{54.2}   \textcolor{darkpastelgreen}{\scriptstyle \vartriangle 137\%} $ &
  $38.0   \textcolor{darkpink}{\scriptstyle \blacktriangledown 30\%} $ &
  $\textbf{64.8}   \textcolor{darkpastelgreen}{\scriptstyle \vartriangle 183\%} $ &
  $36.1   \textcolor{darkpink}{\scriptstyle \blacktriangledown 44\%} $ &
  $-52\%$ \\
  Perpl. & \textbf{Phi1.5, GB} &
  sent. &
  $ 52.5 \textcolor{darkpastelgreen}{\scriptstyle \vartriangle 132 \%}$ &
  $ \textbf{52.0} \textcolor{darkpink}{\scriptstyle \blacktriangledown 1 \%}$ &
  $ \underline{60.4} \textcolor{darkpastelgreen}{\scriptstyle \vartriangle 167 \%}$ &
  $ \underline{24.1} \textcolor{darkpink}{\scriptstyle \blacktriangledown 60 \%}$ &
  $ 52.5 \textcolor{darkpastelgreen}{\scriptstyle \vartriangle 132 \%}$ &
  $ \textbf{45.7} \textcolor{darkpink}{\scriptstyle \blacktriangledown 12 \%}$ &
  $ 48.7 \textcolor{darkpastelgreen}{\scriptstyle \vartriangle 115 \%}$ &
  $ \textbf{56.1} \textcolor{darkpastelgreen}{\scriptstyle \blacktriangle 15 \%}$ &
  $-\textbf{15}\%$ \\
Perpl. & Phi1.5,
  LR &
  sent. &
  $ 48.7 \textcolor{darkpastelgreen}{\scriptstyle \vartriangle 115 \%}$ &
  $ 41.2 \textcolor{darkpink}{\scriptstyle \blacktriangledown 15 \%}$ &
  $ \textbf{60.9} \textcolor{darkpastelgreen}{\scriptstyle \vartriangle 165 \%}$ &
  $ 21.1 \textcolor{darkpink}{\scriptstyle \blacktriangledown 65 \%}$ &
  $ 50.8 \textcolor{darkpastelgreen}{\scriptstyle \vartriangle 125 \%}$ &
  $ \underline{45.2} \textcolor{darkpink}{\scriptstyle \blacktriangledown 11 \%}$ &
  $ 47.1 \textcolor{darkpastelgreen}{\scriptstyle \vartriangle 108 \%}$ &
  $ 51.5 \textcolor{darkpastelgreen}{\scriptstyle \blacktriangle 9 \%}$ &
  $-21\%$ \\
  Perpl. & Phi2, GB &
  sent. &
  $ 50.1 \textcolor{darkpastelgreen}{\scriptstyle \vartriangle 121 \%}$ &
  $ \underline{46.1} \textcolor{darkpink}{\scriptstyle \blacktriangledown 8 \%}$ &
  $ 55.6 \textcolor{darkpastelgreen}{\scriptstyle \vartriangle 146 \%}$ &
  $ 22.8 \textcolor{darkpink}{\scriptstyle \blacktriangledown 41 \%}$ &
  $ 55.0 \textcolor{darkpastelgreen}{\scriptstyle \vartriangle 143 \%}$ &
  $ 42.1 \textcolor{darkpink}{\scriptstyle \blacktriangledown 23 \%}$ &
  $ 48.1 \textcolor{darkpastelgreen}{\scriptstyle \vartriangle 113 \%}$ &
  $ \underline{53.9} \textcolor{darkpastelgreen}{\scriptstyle \blacktriangle 12 \%}$ &
  $-\textbf{15}\%$ \\
Perpl. & Phi2,
  LR &
  sent. &
  $ 49.6 \textcolor{darkpastelgreen}{\scriptstyle \vartriangle 119 \%}$ &
  $ 42.1 \textcolor{darkpink}{\scriptstyle \blacktriangledown 15 \%}$ &
  $ 56.4 \textcolor{darkpastelgreen}{\scriptstyle \vartriangle 149 \%}$ &
  $ \textbf{24.4} \textcolor{darkpink}{\scriptstyle \blacktriangledown 43 \%}$ &
  $ 50.5 \textcolor{darkpastelgreen}{\scriptstyle \vartriangle 123 \%}$ &
  $ 43.4 \textcolor{darkpink}{\scriptstyle \blacktriangledown 14 \%}$ &
  $ 48.5 \textcolor{darkpastelgreen}{\scriptstyle \vartriangle 114 \%}$ &
  $ 52.0 \textcolor{darkpastelgreen}{\scriptstyle \blacktriangle 7 \%}$ &
  $-\underline{16}\%$ \\
  PHD &
  TS multi &
  100 tkn &
  $20.3   \textcolor{darkpink}{\scriptstyle \triangledown 10\%}$ &
  $13.7   \textcolor{darkpink}{\scriptstyle \blacktriangledown 32\%}$ &
  $19.2   \textcolor{darkpink}{\scriptstyle \triangledown 15\%}$ &
  $19.5   \textcolor{darkpastelgreen}{\scriptstyle \blacktriangle 02\%}$ &
  $20.9   \textcolor{darkpink}{\scriptstyle \triangledown 08\%}$ &
  $17.2   \textcolor{darkpink}{\scriptstyle \blacktriangledown 18\%}$ &
  $21.2   \textcolor{darkpink}{\scriptstyle \triangledown 06\%}$ &
  $17.6   \textcolor{darkpink}{\scriptstyle \blacktriangledown 17\%}$ &
  $-\underline{16}\%$ \\
TLE &
  TS Binary & %
  20 tkn &
  $25.6   \textcolor{darkpastelgreen}{\scriptstyle \vartriangle 13\%}$ &
  $14.7   \textcolor{darkpink}{\scriptstyle \blacktriangledown 42\%}$ &
  $16.5   \textcolor{darkpink}{\scriptstyle \triangledown 27\%}$ &
  $16.3   \textcolor{darkpink}{\scriptstyle \blacktriangledown 01\%}$ &
  $25.0   \textcolor{darkpastelgreen}{\scriptstyle \vartriangle 11\%}$ &
  $17.1   \textcolor{darkpink}{\scriptstyle \blacktriangledown 32\%}$ &
  $22.1   \textcolor{darkpink}{\scriptstyle \triangledown 02\%}$ &
  $11.1   \textcolor{darkpink}{\scriptstyle \blacktriangledown 50\%}$ &
  $-31\%$ \\
\midrule
\multicolumn{2}{c}{Best combination}  &
  ~ &
  $63.0   \textcolor{darkpastelgreen}{\scriptstyle \vartriangle 179\%}$ &
  $42.2   \textcolor{darkpink}{\scriptstyle \blacktriangledown 33\%}$ &
  $67.6   \textcolor{darkpink}{\scriptstyle \vartriangle 199\%}$ &
  $20.0   \textcolor{darkpink}{\scriptstyle \blacktriangledown 70\%}$ &
  $60.0   \textcolor{darkpastelgreen}{\scriptstyle \vartriangle 165\%}$ &
  $47.5   \textcolor{darkpink}{\scriptstyle \blacktriangledown 25\%}$ &
  $60.9  \textcolor{darkpastelgreen}{\scriptstyle \vartriangle 169\%}$ &
  $56.4   \textcolor{darkpink}{\scriptstyle \blacktriangledown 07\%}$ &
  $-42\%$ \\
\midrule
Len & GB &
  sent. &
  $28.1   \textcolor{darkpastelgreen}{\scriptstyle \vartriangle 24\%}$ &
  $11.8   \textcolor{darkpink}{\scriptstyle \blacktriangledown 58\%}$ &
  $21.1   \textcolor{darkpink}{\scriptstyle \triangledown 07\%}$ &
  $15.5   \textcolor{darkpink}{\scriptstyle \blacktriangledown 26\%}$ &
  $30.4   \textcolor{darkpastelgreen}{\scriptstyle \vartriangle 34\%}$ &
  $18.4   \textcolor{darkpink}{\scriptstyle \blacktriangledown 39\%}$ &
  $32.3   \textcolor{darkpastelgreen}{\scriptstyle \vartriangle 43\%}$ &
  $15.8   \textcolor{darkpink}{\scriptstyle \blacktriangledown 51\%}$ &
  $-44\%$ \\
Len &
  LR &
  sent. &
  $19.6   \textcolor{darkpink}{\scriptstyle \triangledown 14\%}$ &
  $10.8   \textcolor{darkpink}{\scriptstyle \blacktriangledown 45\%}$ &
  $17.0   \textcolor{darkpink}{\scriptstyle \triangledown 25\%}$ &
  $12.9   \textcolor{darkpink}{\scriptstyle \blacktriangledown 24\%}$ &
  $22.2   \textcolor{darkpink}{\scriptstyle \triangledown 02\%}$ &
  $09.1   \textcolor{darkpink}{\scriptstyle \blacktriangledown 59\%}$ &
  $22.9   \textcolor{darkpastelgreen}{\scriptstyle \vartriangle 01\%}$ &
  $09.1   \textcolor{darkpink}{\scriptstyle \blacktriangledown 60\%}$ &
  $-47\%$ \\
\multicolumn{2}{l}{Majority} &
  --- &
  \multicolumn{2}{c|}{$15.4 \textcolor{darkpink}{\scriptstyle \triangledown 32\%}$} &
  \multicolumn{2}{c|}{$13.0 \textcolor{darkpink}{\scriptstyle   \triangledown 43\%}$} &
  \multicolumn{2}{c|}{$15.9 \textcolor{darkpink}{\scriptstyle   \triangledown 30\%}$} &
  \multicolumn{2}{c|}{$17.7   \textcolor{darkpink}{\scriptstyle \triangledown 22\%}$} &
  --- \\ %
\multicolumn{2}{r}{Approx. human} &
  global &
  \multicolumn{2}{c|}{$22.62$} &
  \multicolumn{2}{c|}{$22.62$} &
  \multicolumn{2}{c|}{$22.62$} &
  \multicolumn{2}{c|}{$22.62$} &
  --- \\ \bottomrule
\end{tabular}%
}
\end{table*}

\textbf{Cross-domain generalization}.
Supervised ATD methods with fine-tuning such as RoBERTa are more sensitive to spurious correlations in a dataset and often demonstrate poor cross-domain transfer, especially compared to topology-based approaches \citep{tulchinskii2023intrinsic}. Table~\ref{tab:cross_roft} reports the results of cross-domain transfer between four text topics in the \emph{RoFT-chatgpt} dataset.
We report in-domain (IN) accuracy on domains seen during training and out-of-domain (OUT) accuracy on the unseen domain 
corresponding to this column; MSE scores are given in Table~\ref{tab:cross_roft_mse} (Appendix~\ref{sec:appendix_crossdomain}).
Each model was trained on three domains and tested on the fourth, unseen domain; we used $60\%$ of these subsets mixed together as the training set, $20\%$ as the validation set, and $20\%$ as the test set for in-domain evaluation.
Table~\ref{tab:cross_roft} shows that RoBERTa's performance drops for all subsets very significantly, 
while perplexity-based classifiers demonstrate excellent cross-domain generalization for \emph{Presidential Speeches} and \emph{Short Stories}; for the \emph{Recipes} domain, TTS classifiers prove to be the most stable, and the \emph{New York Times} subset yields a large generalization gap for all classifiers. 
This means that every type of classifier can handle its own set of spurious features well, and no classifier is universally better than the others. Aggregation of different features can improve the results; for example, a classifier that returns the maximum of RoBERTa and perplexity-based predictions yields OOD scores on average 3\% higher than the best result in Table~\ref{tab:cross_roft} with
comparable ID performance.
As for classification accuracy, surprisingly, the perplexity-based classifier outperforms a fully fine-tuned RoBERTa on all subsets. Moreover, for \emph{Recipes} the multilabel topological time series method places second; this happens because TDA-based methods are extremely stable under domain shift
(despite being worse than others in absolute values on this dataset); see Section~\ref{sec:analysis} for further discussion. 
In general, we conclude that perplexity-based classifiers outperform RoBERTa-based in cross-domain settings by a large margin.
This is remarkable since
the former is a simple classifier trained on ten features extracted by a LM with frozen weights, while the latter involves full LM fine-tuning.
As for the length-based baseline, for in-domain data average sentence length provides a strong signal, leading to accuracy of 20--32\% 
and even outperforming topological methods. But cross-domain generalization fails, which means that we should prefer classifiers that ignore this feature to achieve good generalization (see also Section~\ref{sec:analysis}).

\textbf{Cross-model generalization}.
We tune our classifiers on generation results produced by one model and test the performance for all other models. In general, this task is harder for all considered classifiers: there are models for which prediction accuracy drops down to virtually zero values (detailed experimental results are shown in Appendix~\ref{sec:crossmodel_appendix}: Tables~\ref{tbl:results-1-good}, \ref{tbl:results-1}, \ref{tbl:results-2}). But we observe an interesting result for the perplexity-based classifier: it achieves good generalization when transferring to very large models such as \emph{GPT3-davinci} and GPT2-XL, while for other models the generalization is poor. In general, 
the strongest classifiers fail to generalize to \emph{simpler} models; e.g., RoBERTa-based classifiers underperform on the generalization to GPT-2 and baseline; the baseline is also the hardest subset for perplexity-based classifiers (see Section~\ref{sec:analysis} for a discussion).

\begin{table*}[!t]
  \caption{Cross-model transfer, original RoFT. Models are trained on all parts except one and tested in-domain (ID) on the same parts and out-of-domain (OOD) on the remaining part.}\label{tbl:results-1-good}
  \centering
  \resizebox{\linewidth}{!}{
  \begin{tabular}{l|rr|rr|rr|rr|rr|rr}\toprule
    \textbf{Model} & \multicolumn{2}{c|}{\textbf{GPT2-XL}} & \multicolumn{2}{c|}{\textbf{GPT2}} & \multicolumn{2}{c|}{\textbf{davinci}}& \multicolumn{2}{c|}{\textbf{ctrl-Politics}} & \multicolumn{2}{c|}{\textbf{ctrl-nocode}} & \multicolumn{2}{c}{\textbf{tuned}}  \\
    & \footnotesize{ID}  & \footnotesize{OOD}  & \footnotesize{ID}  & \footnotesize{OOD} & \footnotesize{ID}  & \footnotesize{OOD} & \footnotesize{ID}  & \footnotesize{OOD}  & \footnotesize{ID}  & \footnotesize{OOD} & \footnotesize{ID}  & \footnotesize{OOD} \\
    \midrule
    RoBERTa + SEP & \textbf{46.4}  & \textbf{40.9}  & \textbf{46.0}  & 08.8  & \textbf{63.3}  & \textbf{19.7} & \textbf{49.4}  & \textbf{59.1}  & \textbf{50.2} & \textbf{60.6} & \textbf{49.6}  & \textbf{23.9} \\
    RoBERTa & \underline{\underline{46.7}} & \underline{\underline{32.5}} & \underline{\underline{40.3}} & 07.0 & \underline{\underline{57.2}} & 14.5 & \underline{\underline{47.5}} & \underline{\underline{44.0}} & \underline{\underline{46.1}} & \underline{\underline{54.6}} & \underline{\underline{46.0}} & \underline{\underline{20.1}} \\
    \emph{Phi-1.5 Perpl.} + GB & 26.0 & \underline{27.0} & 31.5 & \underline{12.2} & 30.5 & \underline{15.3} & 30.8 & \underline{23.9} &   30.3 & 19.2 & 34.7 & \underline{12.9} \\
    \emph{Phi-2 Perpl.} + GB & \underline{34.1} & 26.4  & \underline{32.0}  & \underline{\underline{17.6}} & \underline{35.9} & \underline{\underline{16.2}}  & \underline{34.1} & 19.5 &   \underline{34.3} &  \underline{23.2}  & \underline{37.2} & 12.4 \\
    PHD + TS ML &31.8 &  04.0 & 25.6 & 04.5 & 31.4 & 02.1 & 25.7 & 08.2  &  25.2 & 11.1 & 21.4 & 12.6  \\
    TLE + TS Binary & 14.2 & 03.2 & 12.4 & 07.8 & 14.5 & 01.0 & 11.0 & 05.0 & 11.5 & 04.0 & 14.5 & 07.0  \\
    \emph{Length} + GB & 21.7 & 04.8 & 18.3 & 01.2 & 19.2 & 06.3 & 17.3 & 03.8 & 18.1 & 00.0 & 21.6 & 00.0  \\
    Human Baseline &  22.5 &  17.2 &  22.6 &  \textbf{22.5} &  24.7 &  14.1 &  22.6 &  21.6 &  22.6 &  23.9 &  21.5 &  25.9 \\\bottomrule
  \end{tabular}
  }
\end{table*}

\section{Discussion and analysis of the results}\label{sec:analysis}

\begin{figure*}[!t]
    \centering
    \includegraphics[width=\textwidth]{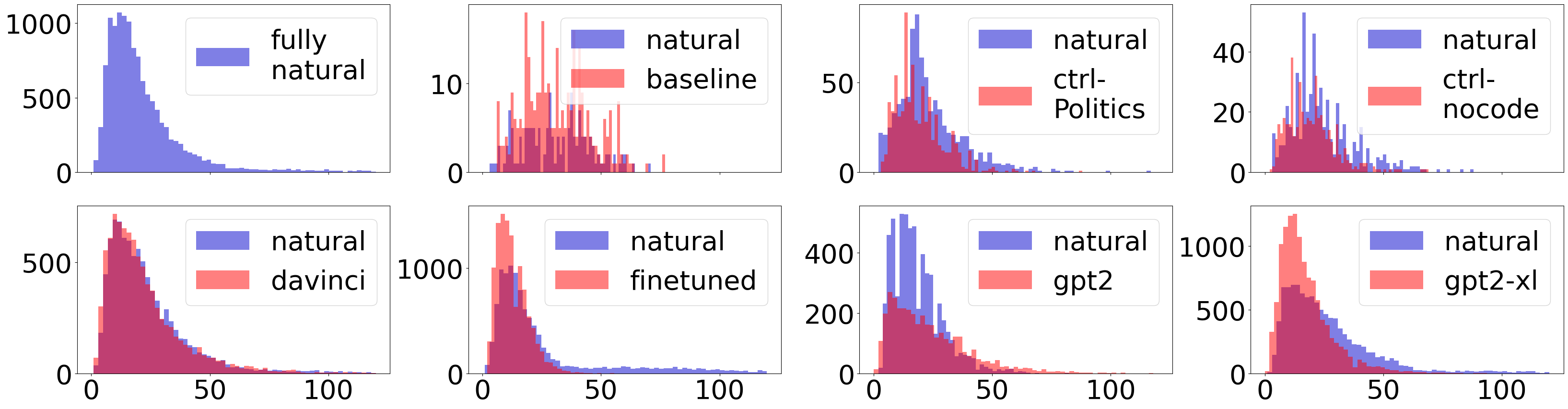}
    \caption{Sentence length distributions in RoBERTA tokens, original RoFT, by model}
    \label{fig:roft-by-model}

    \setlength{\tabcolsep}{20pt}
    \begin{tabular}{cc}
    \includegraphics[height=0.35\textwidth]{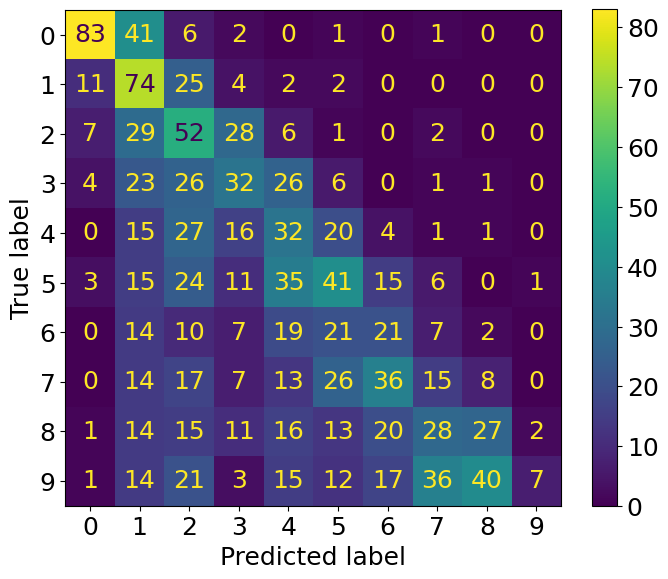}    &
    \includegraphics[height=0.35\textwidth]{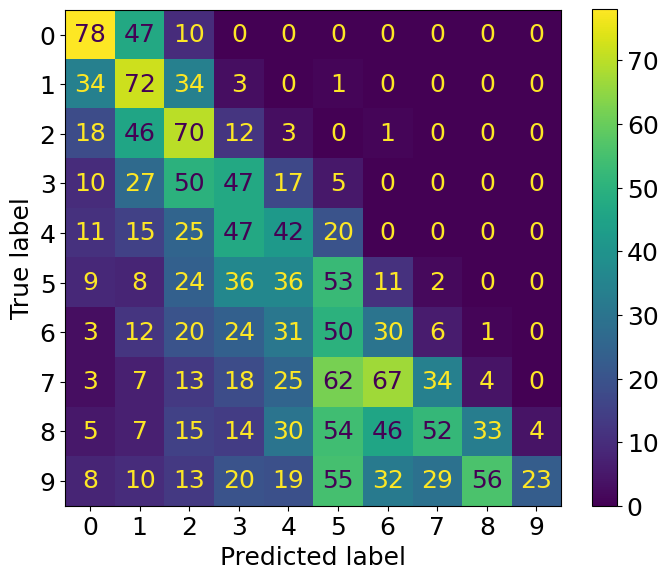} \\
    \includegraphics[height=0.35\textwidth]{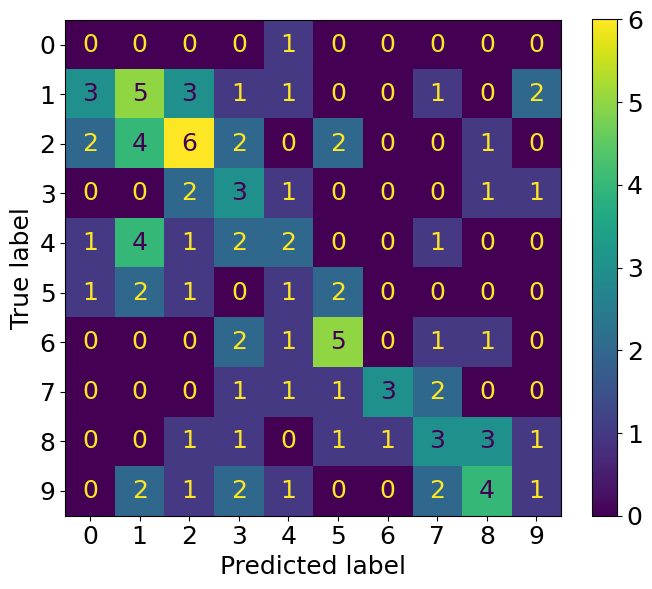} &
    \includegraphics[height=0.35\textwidth]{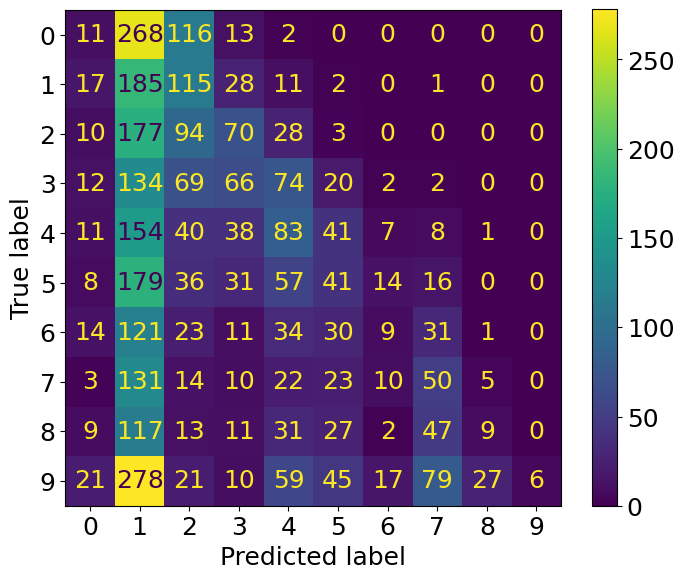}
    \end{tabular}
    \caption{Confusion matrices of RoBERTa predictions on the four domains in the \emph{RoFT-chatgpt} dataset: New York Times, Short Stories, Presidental Speeches, Recipes (left to right).}
    \label{fig:conf_matrix_roberta}
\end{figure*}

\textbf{Detection models}. 
We have found that although RoBERTa-based classifiers demonstrate excellent results for in-domain classification, they lose to perplexity-based methods when tested on texts with new styles and topics not present in the training set. For cross-model generalization, all methods demonstrate insufficient generalization abilities, especially when tested on generations of much larger or much smaller models. This suggests that AI content detectors can be fooled by either too good or \textit{too bad} artificial texts.
For perplexity-based classifiers, we have found that the size of the base model is not the main requirement for a good and robust detector. The training dataset for perplexity estimation is more important;
e.g., ChatGPT generations were best detected by a much smaller model pretrained on synthetic data generated by the same model family. The above observations suggest that pre-training of the perplexity estimators on a large amount of synthetic data from different generators may become the best choice for AI-generated content detection. We leave it for future research.
Topological classifiers demonstrated high robustness over domain transfer, but not very high overall accuracy. 
Geometric properties of the embeddings for both RoFT and \emph{RoFT-chatgpt}, however, show the difference between PHD distributions of real and fake RoFT text for different models (Fig.~\ref{fig:hist_part_models}) and topics (Figs.~\ref{fig:hist_part_topics}, \ref{fig:hist_part_topics_chat} in Appendix~\ref{sec:dataset_appendix}). TLE dimension distributions for sentences in human-written and AI-generated parts of the texts are different as well (Figs.~\ref{fig:hist_sent_models}, \ref{fig:hist_sent_topics}, \ref{fig:hist_sent_topics_chat}, Appendix~\ref{sec:dataset_appendix}). This means that topological classifiers may help in robust artificial content detection in some particular domains and models.

\textbf{Data properties}. Next, we summarize the observed properties of the data that influence detection quality and create difficulties for different types of artificial content detectors.
\begin{enumerate}[wide]
\item First, the \emph{length of sentences} seems to play an important role, deceiving our classifiers. There is indeed a significant difference between distributions of sentence lengths written by humans and generated by LLMs (see Fig.~\ref{fig:roft-by-model} and Figs.~\ref{fig:roft-by-topic},~\ref{fig:roft-chatgpt-by-topic} in Appendix~\ref{sec:dataset_appendix}). 
This is supported by our experiments with the length-based classifier, which sometimes outperforms other methods in terms of accuracy (Tables ~\ref{tab:results} and \ref{tab:cross_roft}), but fails to generalize. 
\item 
\emph{Label distributions} (see Figs.~\ref{fig:roft-labels-by-model} and~\ref{fig:roft-labels-by-topic} in Appendix~\ref{sec:dataset_appendix})
vary significantly across models.
GPT-2 is especially different from natural texts, and the behavior of our models in cross-model transfer to GPT-2
(Appendix~\ref{sec:crossmodel_appendix}), suggests that the only model stable under the label distribution shift is perplexity-based regression. Although its accuracy numbers are low, it outperforms the human level in terms of MSE even in out-of-domain evaluation.
\item Special \emph{text structure} can hurt ATD: e.g., recipes often contain numbered lists, and the first sentence tends to be comparatively long, and as a result the \emph{Recipes} topic is hard for all classifiers to generalize to.
The RoBERTa classifier loses the most here, so we investigated its confusion matrices; Fig.~\ref{fig:conf_matrix_roberta} shows 
that RoBERTa tends to classify all examples in the \emph{Recipes} topic as fully generated.
This kind of text structure can be viewed as an \textit{adversarial example} for AI-generated texts, and indeed, 
ChatGPT is known for its tendency to generate structured output with bullet lists.
\item \emph{Semantic and grammar inconsistencies}. Surprisingly, cross-model generalization of RoBERTa-based and perplexity-based classifiers fails not only for the \textit{best} OOD generator (davinci), but also for the \textit{worst} generators: small-size GPT-2 and baseline. 
Perplexity distributions for texts generated by \emph{baseline} or GPT-2 are very different from the distributions of texts generated by other models (see Fig.~\ref{fig:roft-dist-9th-by-model}). This might explain the poor performance of decision trees and linear classifiers when these generators are excluded from the training set.
On the other hand, both generators are among the easiest for human raters, because of the high level of semantic inconsistencies and grammar issues. It seems that the supervised classifier discovers non-regularities in the text as a feature of human-generated texts; indeed, they present in our training set as a part of first-person narratives, especially in the Short Stories dataset. 
\item \emph{Discourse structure}. Finally, analysing the poor generalization to narrative domains, we discover a subtle property of human-generated texts that influences boundary detection, namely the underlying discourse structure of the story that can be measured as peaks of perplexity along the story (Fig.~\ref{fig:perp_by_sent_by_label}, Appendix~\ref{sec:dataset_appendix}). Human-generated texts tend to have 2-3 such peaks, e.g. a peak in the middle of the story corresponding to a plot twist or change of narrative focus. This leads to the difficulties in perplexity-based boundary detection: the perplexity drop in the beginning of the fake text part is confused with the same drop due to the story discourse change. 
Strong perplexity-based classifiers can handle this issue due to the \textit{std} feature reflecting token-level perplexity deviation
(Fig.~\ref{fig:heatmaps}, Appendix~\ref{sec:dataset_appendix}). Fig.~\ref{fig:roft-by-model}b shows that the RoBERTa-based classifier is prone to this discourse-related issue: it often incorrectly predicts the real-fake boundary in the middle of a human-generated story (5-6th sentence).
\end{enumerate}
A full analysis of feature distributions across domains and models is given in the Appendix.

\section{Conclusion}\label{sec:concl}

In this work, we address the task of boundary detection between human-written and generated parts in texts that combine both. We believe that this setting is increasingly important in real world applications and is a natural setting for recognizing artificial text since it is often mixed with and prompted by human text. 
We have considered the RoFT dataset, presented its modification \emph{RoFT-chatgpt} generated with a more modern LLM, and investigated the performance of features that were useful for artificial text detection in previous works. In particular, we have shown that LLM fine-tuning works reasonably well for this task but tends to overfit to spurious features in the data, which leads to generalization failures in some settings. 
On the other hand, perplexity-based and topological features provide a signal that can help in these situations. We have demonstrated that perplexity features are the best overall on balance between accuracy, generalization, and training complexity.%

We analyze the base models for perplexity estimation and conclude, that reasonably small models work well, but pre-training data is important. Including synthetic data in the pre-training dataset improves the performance of the detector. As for the limitations of our approach, boundary detection is very hard in cross-domain and cross-model settings, both for short texts such as RoFT (due to lack of information) and longer texts (due to a large space of possibilities). Therefore, it is no wonder that none of the methods have achieved a really high quality in this setting, and the results suggest a large room for improvement. Besides, all methods we considered were based on Transformers with relatively small context window size, which limits the transferability of the proposed approaches onto longer text samples. 
Finally, our analysis has uncovered gaps in current approaches and discovered the more difficult aspects of the task, which we plan to address in future research.

\subsubsection*{Acknowledgments}
The work of Sergey Nikolenko was performed at the Saint Petersburg Leonhard Euler International Mathematical Institute and supported by the Ministry of Science and Higher Education of the Russian Federation (agreement no.~075-15-2022-289 dated 06/04/2022).

\bibliography{main}
\bibliographystyle{colm2024_conference}

\appendix

\section{Intrinsic Dimension Estimation Methods}
\label{sec:id_appendix}

According to the manifold hypothesis \citep{Narayanan2010SampleCO}, the data $X$ lies on a low-dimensional submanifold: $X \subseteq M^{n} \subseteq {R}^{d}$, where $d$ is the extrinsic dimension and $n$ is the intrinsic dimension (ID). The geometric and topological properties of the manifold $M$ are of particular interest. There are various methods for estimating the ID that can be divided into global and local methods. %

For the tight local intrinsic dimension estimator (TLE) proposed by \citet{amsaleg:hal-02125331}, we use the neighborhood center point $x$, a set of neighborhood samples $V$ and a specially defined distance between points in a sufficiently small neighborhood of $x$:
\begin{equation}
    d_{x,r}(q,v) = \frac{r(v-q)\cdot (v - q)}{2 (x - q)\cdot (v-q)},
\end{equation}
where $r$ is the radius of the neighbourhood. For every three points $x$, $v$, $w$ we can compute 
$$M(x,v,w) = \ln{\frac{d_{x,r}(v,w)}{r}} +  \ln{\frac{d_{x,r}(2x-v,w)}{r}}.$$
If $V_* = V \cup \{x\}$, then the intrinsic dimension can be found by averaging the estimates for all points $x$, as defined by the following formula:

\begin{equation}
\begin{split}
    \hat{m}_r (x) = - \left (       \frac{1}{|V_*|^2}  \sum_{v, w \in V_*,\ v \neq w}  M(x,v,w)   \right)^{-1}
\end{split}
\end{equation}

Applied algebraic topology provides effective tools for analyzing the topological structure of data. The theoretical foundations of topological data analysis (TDA) have been described in detail by, e.g., \citet{barannikov1994framed} and \citet{carlsson2020persistent}. TDA allows us to consider a dataset $X \subseteq {R}^{d}$ from the topological point of view. In order to move from point clouds $X$ to topological spaces, it is necessary to approximate the data by a simplicial complex $R$. In our research, we use the Vietoris-Rips complex $R(X; t)$. The method of constructing the complex $R$ is as follows: simplexes are formed by subsets of points from X whose pairwise distances do not exceed $t$ (a scaling parameter). An increasing sequence of simplicial complexes is called a filtration:
$\{R_t\}_{(t  \geq 0)}= R_{t_0} \subseteq R_{t_1} \subseteq...\subseteq R_s$.

Homology groups $H_{i}(R)$ are a topological invariant that expresses the properties of a topological space $R$. We use $\beta_i(R)=\dim H_i(R)$, which is known as the $i$th \emph{Betti number}, a topological feature equal to the dimension of the homology group; for $i=0,1,2$ the Betti number corresponds to the number of connectivity components, cycles, and cavities respectively.

Topological features appear and disappear at different values of $t$, which leads to the next core concept in TDA: the \emph{barcode}. It summarises the dynamics of topological features in the filtration process.
A \emph{bar} is the lifetime of the $n$th homology feature $I_n = t^{\mathrm{birth}}_n - t^{\mathrm{death}}_n$. A long bar in the barcode means that the data contains a fairly persistent and informative topological feature.

\citet{schweinhart2020fractal} introduced the persistent homological fractal dimension (PHD) that generalized \citet{steele1988growth} for higher dimensions of the homology group and used the topological properties of the point cloud. The PHD has already been proven to be useful in the study of the properties of deep learning models \citep{birdal2021intrinsic, magai2023deep}.

Let us denote the power-weighted sum of $N$ bars for the $i$th degree of homology as follows:
\begin{equation}\label{eq:trigrule} 
E_\alpha^i(X) = \sum_{i=1}^{N} I^\alpha_i.
\end{equation}

It is interesting to note that $E_1^0(X_n)$ is equal to the length of the Euclidean minimum spanning tree (MST) of $X_{n} \subseteq {R}^{d}$ \citep{skraba2017randomly}.

Then the persistent homological fractal dimension (PHD) can be defined as follows:
\begin{equation}\label{eq:trigrule2}
\mathrm{PHD}^{i} = \frac{\alpha}{1-\beta},
\end{equation}
where
\begin{equation}\label{eq:trigrule3}
\beta = \lim\limits_{n \to \infty} \sup \frac{\log(\mathop{\mathbb{E}}(E_\alpha^i(x_1,...,x_n)))}{\log(n)},
\end{equation}
and $x_1,\ldots,x_n$ are sampled independently from $X$. That is, $\mathrm{PHD}^i(X)=d$ if $E_\alpha^i(x_1,...,x_n)$ scales as $n^{\frac{d-\alpha}{d}}$ and $\alpha  \geq  0$ (we take $i=0$, $\alpha=1$). Persistent homological fractal dimension can be estimated by analyzing the asymptotical behavior at $n\rightarrow\infty$ of $E_\alpha^i(x_1...x_n)$ for every $i$. In other words, to calculate PHD we must find a power law that shows how $E_\alpha^i(x_1...x_n)$ scales as $n$ increases. See \citet{adams2020fractal} and \citet{schweinhart2020fractal} for more details.

\begin{figure*}[!t]
     \centering
     \begin{subfigure}[c]{0.55\textwidth}
         \centering
         \includegraphics[width=\textwidth]{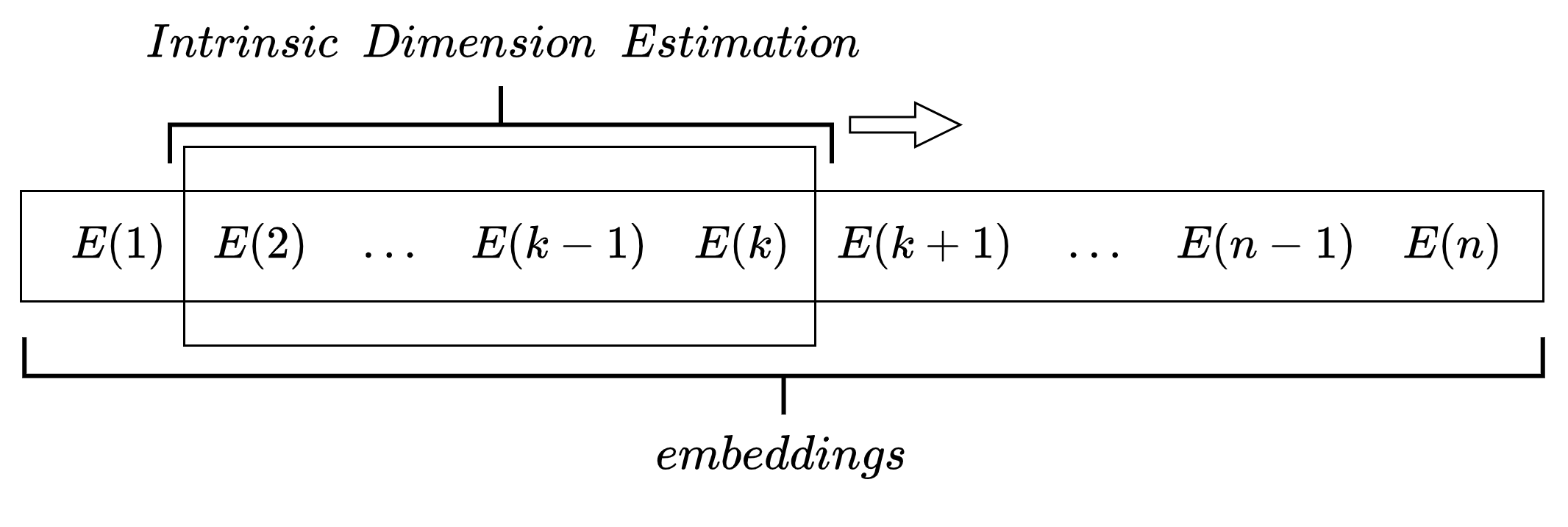}
         \caption{Sliding window technique}
         \label{fig:y equals x}
     \end{subfigure}
     \hfill
     \begin{subfigure}[c]{0.42\textwidth}
         \centering
         \includegraphics[width=\textwidth]{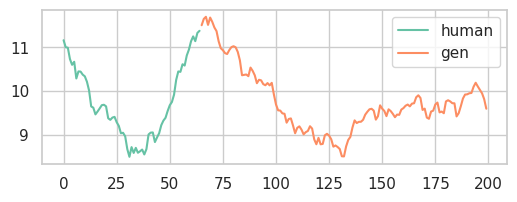}
         \caption{Sample result}
         \label{fig:three sin x}
     \end{subfigure}
    \caption{Multilabel time series classification: (a) estimating the intrinsic dimension of token embeddings in a sliding window, (b) a sample resulting series: green~-- human-written, orange~-- machine-generated tokens.}
    \label{fig:three graphs}
\end{figure*}

\section{gpt-3.5-turbo and gpt-4 zero-shot experiment details}\label{sec:chatgpt}

Below we show the prompt used for gpt-3.5-turbo and gpt-4 zero-shot experiments.

\begin{lstlisting}
You will be provided with a text that contains human-written sentences. Your task is to analyze the text and determine if any of the sentences were generated by an AI system.
You should output a single number between 0 and 9 to indicate where the first AI-generated sentence occurs in the text (if any):
For your output, you should provide a single number between 0 and 9, with the following meaning:
0 - The entire text was generated by an AI system.
1 to 8 - The first AI-generated sentence occurs at that position in the text (1 for the first sentence, 2 for the second sentence, and so on).
9 - The entire text was written by a human with no AI-generated sentences.
Your output should consist solely of this single digit number, with no additional text or explanation:
{content}
\end{lstlisting}

\section{Cross-domain transfer}\label{sec:appendix_crossdomain}

\begin{table*}[!t]
\centering
\setlength{\tabcolsep}{0.15em}
  \caption{
  Accuracy for leave-one-out cross-domain evaluation on \emph{RoFT-chatgpt}, full table.
  }
  \label{tab:cross_roft_acc_full}

\resizebox{\textwidth}{!}{
\begin{tabular}{@{}lll|cc|cc|cc|cc@{}}
\toprule
 &
 \multicolumn{2}{c|}{} &
  \multicolumn{2}{c|}{\textbf{Presidential}} &
  \multicolumn{2}{c|}{\textbf{}} &
  \multicolumn{2}{c|}{\textbf{New York}} &
  \multicolumn{2}{c}{\textbf{Short}} \\ %
\textbf{Pre-} &
 \multicolumn{2}{c|}{} &
  \multicolumn{2}{c|}{\textbf{Speeches}} &
  \multicolumn{2}{c|}{\textbf{Recipes}} &
  \multicolumn{2}{c|}{\textbf{Times}} &
  \multicolumn{2}{c}{\textbf{Stories}} \\ %
\textbf{dictor} &
  \textbf{Model} &
  \textbf{Context} &
  IN $\uparrow$ &
  OUT $\uparrow$ &
  IN $\uparrow$ &
  OUT $\uparrow$ &
  IN $\uparrow$ &
  OUT $\uparrow$ &
  IN $\uparrow$ &
  OUT $\uparrow$  \\ 
  \midrule
  Text &
    RoBERTa SEP&
  global &
  $\textbf{57.3}  $ &
  $31.4  $ &
  $43.2 $ &
  $13.1  $ &
  $\underline{53.2} $ &
  $38.1 $ &
  $\textbf{54.3} $ &
  $28.6  $ \\
Text &
  RoBERTa &
  global &
  $\underline{54.2} $ &
  $40.2  $ &
  $39.7  $ &
  $15.1 $ &
  $\textbf{53.3} $ &
  $34.0 $ &
  $\underline{50.3} $ &
  $27.7 $ \\

  Perpl. & GPT2, GB &
  sentence &
  $36.1  $ &
  $35.3 $ &
  $36.6  $ &
  $19.4 $ &
  $36.9  $ &
  $29.7 $ &
  $32.8  $ &
  $32.9  $  \\
Perpl. & GPT2,
  LR &
  sentence &
  $34.8  $ &
  $32.4 $ &
  $40.6  $ &
  $20.4 $ &
  $36.7  $ &
  $28.9 $ &
  $32.7  $ &
  $35.6  $ \\
Perpl. & GPT2, Regr. (GB) &
  sentence &
  $20.9   $ &
  $22.6   $ &
  $23.8   $ &
  $14.4  $ &
  $18.5   $ &
  $16.1  $ &
  $21.2   $ &
  $21.8 $ \\
  Perpl. & Phi1, GB &
  sentence &
  $ 36.2 $ &
  $ 24.5$ &
  $ 26.5 $ &
  $ 21.6$ &
  $ 39.1 $ &
  $ 23.8$ &
  $ 42.1 $ &
  $ 21.2$ \\
Perpl. & Phi1,
  LR &
  sentence &
  $ 33.6 $ &
  $ 27.4$ &
  $ 25.9 $ &
  $ 26.2$ &
  $ 37.4 $ &
  $ 25.5$ &
  $ 35.5 $ &
  $ 22.9 $ \\
  Perpl. & \textbf{Phi1.5, GB} &
  sentence &
  $ 52.5 $ &
  $ \textbf{52.0} $ &
  $ \underline{60.4} $ &
  $ \underline{24.1}$ &
  $ 52.5 $ &
  $ \textbf{45.7} $ &
  $ 48.7 $ &
  $ \textbf{56.1} $   \\
Perpl. & Phi1.5,
  LR &
  sentence &
  $ 48.7 $ &
  $ 41.2 $ &
  $ \textbf{60.9} $ &
  $ 21.1 $ &
  $ 50.8 $ &
  $ \underline{45.2} $ &
  $ 47.1 $ &
  $ 51.5 $  \\
  Perpl. & Phi2, GB &
  sentence &
  $ 50.1 $ &
  $ \underline{46.1} $ &
  $ 55.6 $ &
  $ 22.8 $ &
  $ 55.0 $ &
  $ 42.1 $ &
  $ 48.1 $ &
  $ \underline{53.9}$  \\
Perpl. & Phi2,
  LR &
  sentence &
  $ 49.6 $ &
  $ 42.1$ &
  $ 56.4 $ &
  $ \textbf{24.4} $ &
  $ 50.5 $ &
  $ 43.4$ &
  $ 48.5 $ &
  $ 52.0 $ \\
  Perpl. & LLaMA2-7B, GB &
  sentence &
  $ 41.4 $&
  $ 25.5$&
  $ 39.0 $&
  $ 17.7$&
  $ 45.2 $&
  $ 24.5$&
  $ 41.1 $&
  $ 27.4 $ \\
Perpl. & LLaMA2-7B,
  LR &
  sentence &
  $ 34.5 $ &
  $ 15.7$ &
  $ 37.9 $ &
  $ 14.8$ &
  $ 37.2 $ &
  $ 23.5$ &
  $ 36.1 $ &
  $ 24.4 $  \\

  PHD &
  TS multilabel &
  100 tokens &
  $20.3 $ &
  $13.7 $ &
  $19.2 $ &
  $19.5 $ &
  $20.9 $ &
  $17.2 $ &
  $21.2 $ &
  $17.6 $ \\
TLE &
  TS Binary & %
  20 tokens &
  $25.6  $ &
  $14.7 $ &
  $16.5  $ &
  $16.3 $ &
  $25.0  $ &
  $17.1 $ &
  $22.1  $ &
  $11.1 $  \\

Length & GB &
  sentence &
  $28.1  $ &
  $11.8 $ &
  $21.1 $ &
  $15.5 $ &
  $30.4  $ &
  $18.4 $ &
  $32.3  $ &
  $15.8 $ \\
Length &
  LogRegr &
  sentence &
  $19.6 $ &
  $10.8 $ &
  $17.0 $ &
  $12.9 $ &
  $22.2 $ &
  $09.1$ &
  $22.9 $ &
  $09.1$  \\
\multicolumn{2}{l}{Majority} &
  --- &
  \multicolumn{2}{c|}{15.4} &
  \multicolumn{2}{c|}{13.0} &
  \multicolumn{2}{c|}{15.9} &
  \multicolumn{2}{c}{17.7}  \\ \midrule
\multicolumn{2}{c}{Approximated   human\*\*} &
  global &
  \multicolumn{2}{c|}{$22.62$} &
  \multicolumn{2}{c|}{$22.62$} &
  \multicolumn{2}{c|}{$22.62$} &
  \multicolumn{2}{c}{$22.62$} \\ \bottomrule
\end{tabular}%
}
\end{table*}

\begin{table*}[!t]
\setlength{\tabcolsep}{0.15em}
  \caption{Mean squared errors from leave-one-out cross-domain evaluation on \emph{RoFT-chatgpt}, full table.
  }\label{tab:cross_roft_mse}

\resizebox{\textwidth}{!}{%
\begin{tabular}{@{}lll|cc|cc|cc|cc@{}}
\toprule
 &
 \multicolumn{2}{c|}{} &
  \multicolumn{2}{c|}{\textbf{Presidential}} &
  \multicolumn{2}{c|}{\textbf{}} &
  \multicolumn{2}{c|}{\textbf{New York}} &
  \multicolumn{2}{c}{\textbf{Short}} \\ %
\textbf{Pre-} &
 \multicolumn{2}{c|}{} &
  \multicolumn{2}{c|}{\textbf{Speeches}} &
  \multicolumn{2}{c|}{\textbf{Recipes}} &
  \multicolumn{2}{c|}{\textbf{Times}} &
  \multicolumn{2}{c}{\textbf{Stories}} \\ %
\textbf{dictor} &
  \textbf{Model} &
  \textbf{Context} &
  IN $\downarrow$ &
  OUT $\downarrow$ &
  IN $\downarrow$ &
  OUT $\downarrow$ &
  IN $\downarrow$ &
  OUT $\downarrow$ &
  IN $\downarrow$ &
  OUT $\downarrow$  \\ 
  \midrule
  Text & RoBERTa SEP &
  global &
  $ 02.6 $ &
  $ 10.6 $ &
  $ 02.6 $ &
  $ 18.3  $ &
  $ 03.4  $ &
  $ 07.9  $ &
  $ 02.3  $ &
  $ 09.0  $ \\
  Text &
  RoBERTa &
  global &
  $02.3  $ &
  $07.5   $ &
  $02.8  $ &
  $13.5  $ &
  $02.9 $ &
  $06.2   $ &
  $02.6  $ &
  $05.5  $ \\
  Perpl. & GPT2, GB &
  sentence &
  $07.3 $ &
  $08.9 $ &
  $07.0 $ &
  $14.6 $ &
  $07.2 $ &
  $09.4 $ &
  $08.5 $ &
  $08.5 $  \\
Perpl. &
  GPT2, LR &
  sentence &
  $08.2   $ &
  $11.5   $ &
  $06.0  $ &
  $16.8  $ &
  $08.7   $ &
  $11.8  $ &
  $09.6   $ &
  $09.3 $ \\
  Perpl. &
  GPT2, Regr. (GB) &
  sentence &
  $04.7  $ &
  $06.6 $ &
  $04.8  $ &
  $09.7  $ &
  $04.9  $ &
  $05.7  $ &
  $05.2  $ &
  $05.0 $ \\

    Perpl. & Phi1, GB &
  sentence &
  $09.3 $ &
  $11.8 $ &
  $10.8 $ &
  $15.1 $ &
  $08.5 $ &
  $14.2 $ &
  $08.7 $ &
  $14.7 $  \\
Perpl. &
  Phi1, LR &
  sentence &
  $09.8   $ &
  $11.9   $ &
  $11.5  $ &
  $12.9  $ &
  $08.9   $ &
  $12.2 $ &
  $09.1  $ &
  $13.5 $ \\

      Perpl. & Phi1.5, GB &
  sentence &
  $04.5 $ &
  $08.5 $ &
  $03.0 $ &
  $11.8 $ &
  $03.9 $ &
  $06.3 $ &
  $05.6 $ &
  $03.4 $  \\
Perpl. &
  Phi1.5, LR &
  sentence &
  $04.5  $ &
  $09.3   $ &
  $03.6  $ &
  $12.2  $ &
  $04.0$ &
  $07.2 $ &
  $05.2  $ &
  $03.0 $ \\

 Perpl. & Phi2, GB &
  sentence &
  $04.3 $ &
  $07.3 $ &
  $03.0 $ &
  $15.5 $ &
  $04.3 $ &
  $06.6 $ &
  $04.6 $ &
  $04.3 $  \\
Perpl. &
  Phi2, LR &
  sentence &
  $04.5  $ &
  $09.9   $ &
  $03.4  $ &
  $13.6  $ &
  $04.3$ &
  $07.5 $ &
  $05.4  $ &
  $03.0 $ \\

   Perpl. & Phi2, GB &
  sentence &
  $04.3 $ &
  $07.3 $ &
  $03.0 $ &
  $15.5  $ &
  $04.3  $ &
  $06.6  $ &
  $04.6  $ &
  $04.3  $  \\
Perpl. &
  Phi2, LR &
  sentence &
  $04.5   $ &
  $09.9    $ &
  $03.4   $ &
  $13.6   $ &
  $04.3 $ &
  $07.5  $ &
  $05.5   $ &
  $03.0  $ \\

  Perpl. & LLAMA-2-7B, GB &
  sentence &
  $05.0 $ &
  $14.0 $ &
  $05.5  $ &
  $13.2  $ &
  $05.3  $ &
  $11.4  $ &
  $07.3  $ &
  $09.7  $  \\
Perpl. &
  LLAMA-2-7B, LR &
  sentence &
  $06.3   $ &
  $18.3    $ &
  $06.0   $ &
  $19.5   $ &
  $06.1 $ &
  $08.8  $ &
  $08.3   $ &
  $05.0  $ \\
  
PHD &
  TS multilabel &
   100 tokens &
  $12.3 $ &
  $14.6 $ &
  $10.7 $ &
  $11.0 $ &
  $14.1 $ &
  $16.8 $ &
  $11.6 $ &
  $11.6 $ \\
TLE &
  TS Binary & %
 20 tokens &
  $12.3 $ &
  $15.6 $ &
  $18.0 $ &
  $15.4 $ &
  $12.0 $ &
  $17.6 $ &
  $17.4 $ &
  $23.9 $ \\
  
Length & GB &
  sentence &
  $12.8  $ &
  $15.9 $ &
  $14.2 $ &
  $17.9 $ &
  $13.7 $ &
  $18.4 $ &
  $12.5 $ &
  $15.3 $ \\
Length &
  LogReg &
  sentence &
  $20.1 $ &
  $20.5   $ &
  $16.7 $ &
  $24.7   $ &
  $17.2 $ &
  $22.1   $ &
  $18.5 $ &
  $22.9   $ \\

\multicolumn{2}{l}{Majority} &
  --- &
  \multicolumn{2}{c|}{$27.5 $} &
  \multicolumn{2}{c|}{$27.4 $} &
  \multicolumn{2}{c|}{$27.9 $} &
  \multicolumn{2}{c}{$28.0 $}  \\ \midrule
\multicolumn{2}{c}{Approximated   human**} &
  global &
  \multicolumn{2}{c}{13.88} &
  \multicolumn{2}{c}{13.88} &
  \multicolumn{2}{c}{13.88} &
  \multicolumn{2}{c}{13.88} 
   \\ \bottomrule
\end{tabular}%
}
\end{table*}

Tables \ref{tab:cross_roft_mse} and \ref{tab:cross_roft_acc_full} supplement Table~\ref{tab:cross_roft} from the main text, reporting the results of cross-domain transfer for our methods on the \emph{RoFT-chatgpt} dataset across four text topics present in the data. We report in-domain and out-of-domain accuracy: the IN column shows results from domains seen during training, while the OUT column reflects the model's ability to detect artificial texts in the unseen domain corresponding to this column. For each model, training was done on three out of the four domains, and the resulting model was tested on the fourth, unseen domain; we used $60\%$ of these subsets, mixed together, as the training set, $20\%$ as the validation set, and $20\%$ as the test set for in-domain evaluation.

\section{Cross-model transfer}
\label{sec:crossmodel_appendix}

\begin{table*}[!t]
  \caption{Original RoFT, cross-model transfer, part 1. The models were trained on all parts of the dataset except one and tested in-domain (ID) on the same parts and out-of-domain (OOD) on the remaining part.}\label{tbl:results-1}
  \centering
  \resizebox{.95\linewidth}{!}{
  \begin{tabular}{ll|rr|rr|rr}\toprule
    \textbf{Model} & \textbf{Metric} & \multicolumn{2}{c|}{\textbf{GPT2-XL}} & \multicolumn{2}{c|}{\textbf{GPT2}} & \multicolumn{2}{c}{\textbf{davinci}} \\
    & & ID  & OOD  & ID  & OOD & ID  & OOD \\
    \midrule
     RoBERTa + SEP & {Acc, \%} & 46.38  & 40.94  & 45.95  & 08.78  & 63.25  & 19.73  \\
    RoBERTa + SEP & {SoftAcc1, \%} & 76.85  & 76.83  & 76.71  & 31.22  & 85.52  & 47.27  \\
     RoBERTa + SEP & {MSE} & 03.90 & 02.92 & 04.17 & 06.77 & 03.04 & 07.59 \\
     \midrule
     RoBERTa & {Acc, \%} & 46.68  & 32.56  & 40.30  & 06.94  & 57.20  & 14.48  \\
    RoBERTa & {SoftAcc1, \%} & 77.30  & 72.07  & 75.52  &  25.31  & 84.61  & 40.49  \\
     RoBERTa & {MSE} & 03.73 & 03.10 & 03.70 & 07.18 & 02.94 & 08.56 \\
     \midrule
      \emph{GPT-2 Perplexity} + GB & {Acc, \%} & 23.00  &  23.43  & 28.12  & 04.08  & 23.75  & 19.78  \\
    \emph{GPT-2 Perplexity} + GB & {SoftAcc1, \%} & 40.35  & 47.90  & 47.96  & 30.61  & 46.03  & 42.46  \\
    \emph{GPT-2 Perplexity} + GB & {MSE} & 15.39 & 10.51 & 12.19 & 15.99 & 11.85 & 15.38 \\
    \midrule
    \emph{GPT-2 Perplexity} + LogReg & {Acc, \%} & 21.27  &  08.86  & 23.67  & 03.47  & 21.43  & 08.31  \\
   \emph{GPT-2 Perplexity} + LogReg & {SoftAcc1, \%} & 33.33  & 22.14  & 39.80  & 27.45  & 35.35  & 25.19  \\
    \emph{GPT-2 Perplexity} + LogReg & {MSE} & 21.52 & 24.48 & 16.99 & 18.98 & 19.71 & 22.06 \\
    \midrule
    \emph{GPT-2 Perplexity} + Regr & {Acc, \%} & 11.68  &  15.78  & 14.56  & 14.18  & 13.91  & 15.30  \\
    \emph{GPT-2 Perplexity} + Regr & {SoftAcc1, \%} & 34.84  & 49.37  & 39.36  & 47.86  & 46.10  & 44.43  \\
    \emph{GPT-2 Perplexity} + Regr & {MSE} & 07.67 & 04.62 & 06.80 & 06.40 & 06.73 & 06.64 \\
    \midrule
    PHD + TS ML & {Acc, \%} & 31.84  &  04.02  & 25.64  & 04.49  & 31.38  & 02.05  \\
    PHD + TS ML & {SoftAcc1, \%} & 56.08  & 14.14  & 44.28  & 35.85  & 53.31  & 13.82  \\
    PHD + TS ML & {MSE} & 11.13 & 28.18 & 16.37 & 10.74 & 11.82 & 27.03 \\
    \midrule
    \emph{TLE} + TS Binary & {Acc, \%} & 14.17  &  03.15  & 12.36  & 07.76  & 14.48  & 00.98  \\
    \emph{TLE} + TS Binary & {SoftAcc1, \%} & 31.14  & 13.10 & 29.19  & 32.14  & 34.01  & 11.58  \\
    \emph{TLE} + TS Binary & {MSE} & 21.58 & 28.20 & 21.36 & 16.35 & 18.12 & 30.45 \\
    \midrule
    \emph{Length} + GB & {Acc, \%} &  21.70 &  04.75 &  18.30 &  01.22 &  19.18 &  06.33 \\
    \emph{Length} + GB & {SoftAcc1, \%} &  36.00 &  09.48 &  33.20 &  22.02 &  36.09 &  22.07 \\
    \emph{Length} + GB & {MSE} & 23.56  & 33.32 & 27.00 & 19.27 &  17.35 &  20.00  \\
    \midrule
    Human Baseline & {Acc, \%} &  22.48 &  17.23 &  22.59 &  22.53 &  24.74 &  14.06 \\
    Human Baseline & {SoftAcc1, \%} &  41.91 &  37.03 &  39.73 &  48.01 &  42.44 &  33.47 \\
    Human Baseline & {MSE} &  13.49 & 14.69  & 14.29 & 09.86 & 14.03 &  12.91 \\\bottomrule
  \end{tabular}
  }
\end{table*}

\begin{table*}[!t]
  \caption{Original RoFT, cross-model transfer, part 2. The models were trained on all parts of the dataset except one and tested in-domain (ID) on the same parts and out-of-domain (OOD) on the remaining part.}\label{tbl:results-2}
  \centering
  \resizebox{\linewidth}{!}{
  \begin{tabular}{ll|rr|rr|rr|rr}\toprule
    \textbf{Model} & \textbf{Metric} & \multicolumn{2}{c|}{\textbf{ctrl-Politics}} & \multicolumn{2}{c|}{\textbf{ctrl-nocode}} & \multicolumn{2}{c}{\textbf{tuned}} & \multicolumn{2}{c}{\textbf{baseline}} \\
    & & ID  & OOD  & ID  & OOD & ID  & OOD & ID  & OOD \\
    \midrule
    RoBERTa + SEP & {Acc, \%} & 49.35  & 59.12  & 50.20  & 60.61  & 49.55  & 23.85  & 51.35  & 06.35  \\
    RoBERTa + SEP & {SoftAcc1, \%} & 78.49  & 89.31  & 80.10  & 84.85  &  81.55  & 56.96  & 79.84  & 15.87  \\
    RoBERTa + SEP & {MSE} & 02.86 & 01.24 & 02.93 & 01.87 & 02.33 & 06.33 & 02.51 & 39.32 \\
    \midrule
    RoBERTa & {Acc, \%} & 47.46  & 44.03  & 46.07  & 54.55  & 45.91  & 20.98  & 47.29  & 04.76  \\
    RoBERTa & {SoftAcc1, \%} & 78.43  & 85.53  & 78.01  & 86.87  & 80.13  & 52.88  & 78.49  & 15.87  \\
    RoBERTa & {MSE} & 02.80 & 01.21 & 02.79 & 01.07 & 02.37 & 06.49 & 02.69 & 36.00 \\
    \midrule
    \emph{GPT-2 Perpl.} + GB & {Acc, \%} &  25.27  &  10.69  & 25.32   &  07.07  &  30.88  & 10.15   & 24.44   & 06.35  \\
    \emph{GPT-2 Perpl.} + GB & {SoftAcc1, \%} & 48.89  & 27.04  & 47.71  & 24.24  & 53.87  & 29.20  & 47.64 & 14.29  \\
    \emph{GPT-2 Perpl.} + GB & {MSE} & 11.81 & 20.96 & 12.27& 22.70 &  11.67 & 17.86 &  12.01 & 39.62\\
    \midrule
    \emph{GPT-2 Perpl.} + LogReg & {Acc, \%} & 24.08   &  07.55  &  21.88  &   07.07  &  28.50  & 08.04   & 24.77   &  03.17  \\
    \emph{GPT-2 Perpl.} + LogReg & {SoftAcc1, \%} & 42.23  & 23.90  & 38.84  &  19.19  & 42.78   & 22.28  & 42.45  & 15.87  \\
    \emph{GPT-2 Perpl.} + LogReg & {MSE} & 15.70 & 23.27 & 16.69 & 25.38 & 17.46 & 24.20 & 15.72  & 40.86 \\
    \midrule
    \emph{GPT-2 Perpl.} + Regr & {Acc, \%} &  14.80  & 13.21   & 13.96   &  14.14  &  15.70  &  11.90  & 13.62   & 03.17  \\
    \emph{GPT-2 Perpl.} + Regr & {SoftAcc1, \%} & 42.06  & 40.25  & 41.89  & 33.33  &  43.08  & 36.03  & 39.25  & 14.29  \\
    \emph{GPT-2 Perpl.} + Regr & {MSE} & 06.40 & 07.78 & 06.55 & 08.31 & 06.81 & 07.10 & 06.84  & 22.86 \\
    \midrule
    {PHD + TS ML} & {Acc, \%} & 25.70  & 08.18  &  25.18  & 11.11  & 21.44  & 12.58  &  23.13  & 03.70  \\
    {PHD + TS ML} & {SoftAcc1, \%} & 47.33  & 32.70  & 47.30  & 39.39  & 35.31  & 26.37  & 45.64  & 05.56  \\
    {PHD + TS ML} & {MSE} & 14.09 & 11.23 & 13.62 & 09.41 & 18.28 & 18.43 & 13.42 & 52.39  \\
    \midrule
    \emph{TLE} + TS binary & {Acc, \%} & 10.98  &  05.03  & 11.53  & 04.04  & 14.51  & 06.56  & 12.39  & 03.18  \\
    \emph{TLE} + TS binary & {SoftAcc1, \%} & 29.08  & 18.87 & 28.21  & 22.22  & 30.21  & 17.26  & 28.21  & 15.88  \\
    \emph{TLE} + TS binary & {MSE} & 20.36 & 23.28 & 20.84 & 21.82 & 21.82 & 26.64 & 20.59 & 25.55 \\
    \midrule
    \emph{Length} + GB & {Acc, \%} &  17.32 &  03.77 &  18.05 &  0.0 &  21.64 &  0.04 &  15.24 & 26.98 \\
    \emph{Length} + GB & {SoftAcc1, \%} &  33.96 &  18.86 &  35.71 &  15.15 &  35.80 &  10.87 & 32.56 & 34.92 \\
    \emph{Length} + GB & {MSE} &  22.48 & 23.87 & 24.16 & 26.53 & 23.59 & 32.78 &  16.25 & 16.0\\
    \midrule
    Human Baseline & {Acc, \%} &  22.60 &  21.6 &  22.57 &  23.94 &  21.46 &  25.90 &  22.41 & 46.15 \\
    Human Baseline & {SoftAcc1, \%} &  40.61 &  41.6 &  40.59 &  45.07 &  39.17 &  44.92  &  40.46 & 63.46 \\
    Human Baseline & {MSE} & 13.87 & 10.57 & 13.87 & 07.70 & 13.92 & 13.48 & 13.82 & 11.51 \\
    \bottomrule
  \end{tabular}
  }\vspace{.4cm}
\end{table*}

Tables~\ref{tbl:results-1} and~\ref{tbl:results-2} show our experimental results on cross-model transfer for all considered text generation models. The artificial text boundary detection models were trained on the parts of the dataset generated by all language models except one, which is held out for cross-model testing, and tested in-domain (ID) on the same parts and out-of-domain (OOD) on the remaining part generated by the held-out model.

\section{Additional Dataset Analysis}
\label{sec:dataset_appendix}

\begin{figure*}[!t]
    \centering
    \includegraphics[width=\textwidth]{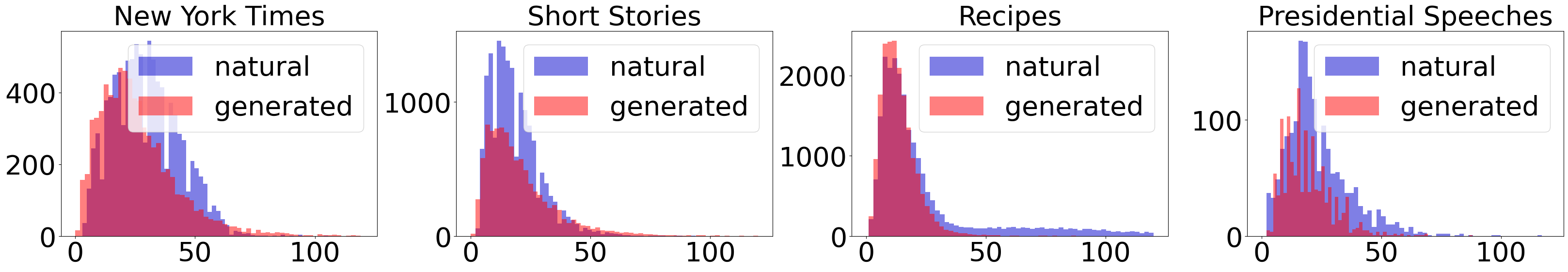}
    \caption{Sentence length distributions in RoBERTA tokens, original RoFT, by topic}
    \label{fig:roft-by-topic}
    \vspace{.6cm}
    
    \centering
    \includegraphics[width=\textwidth]{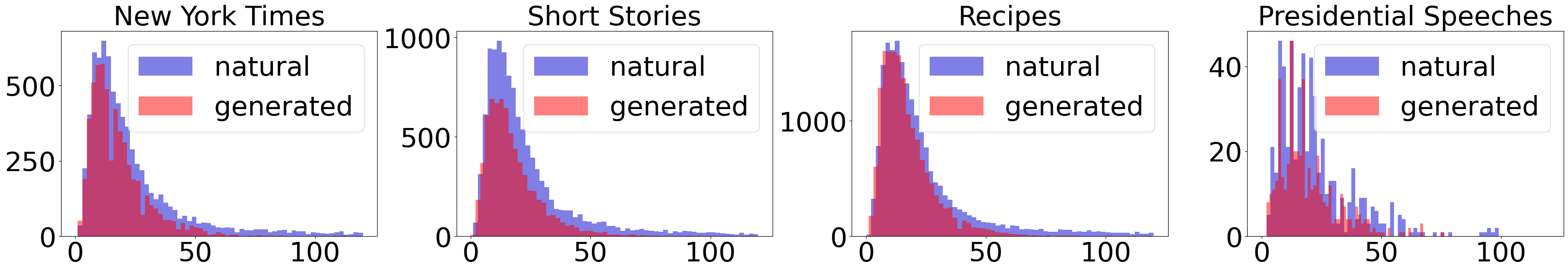}
    \caption{Sentence length distributions in RoBERTA tokens, RoFT-chatgpt, by topic}
    \label{fig:roft-chatgpt-by-topic}
\end{figure*}

\begin{figure*}[!t]
    \centering
    \includegraphics[width=.92\textwidth]{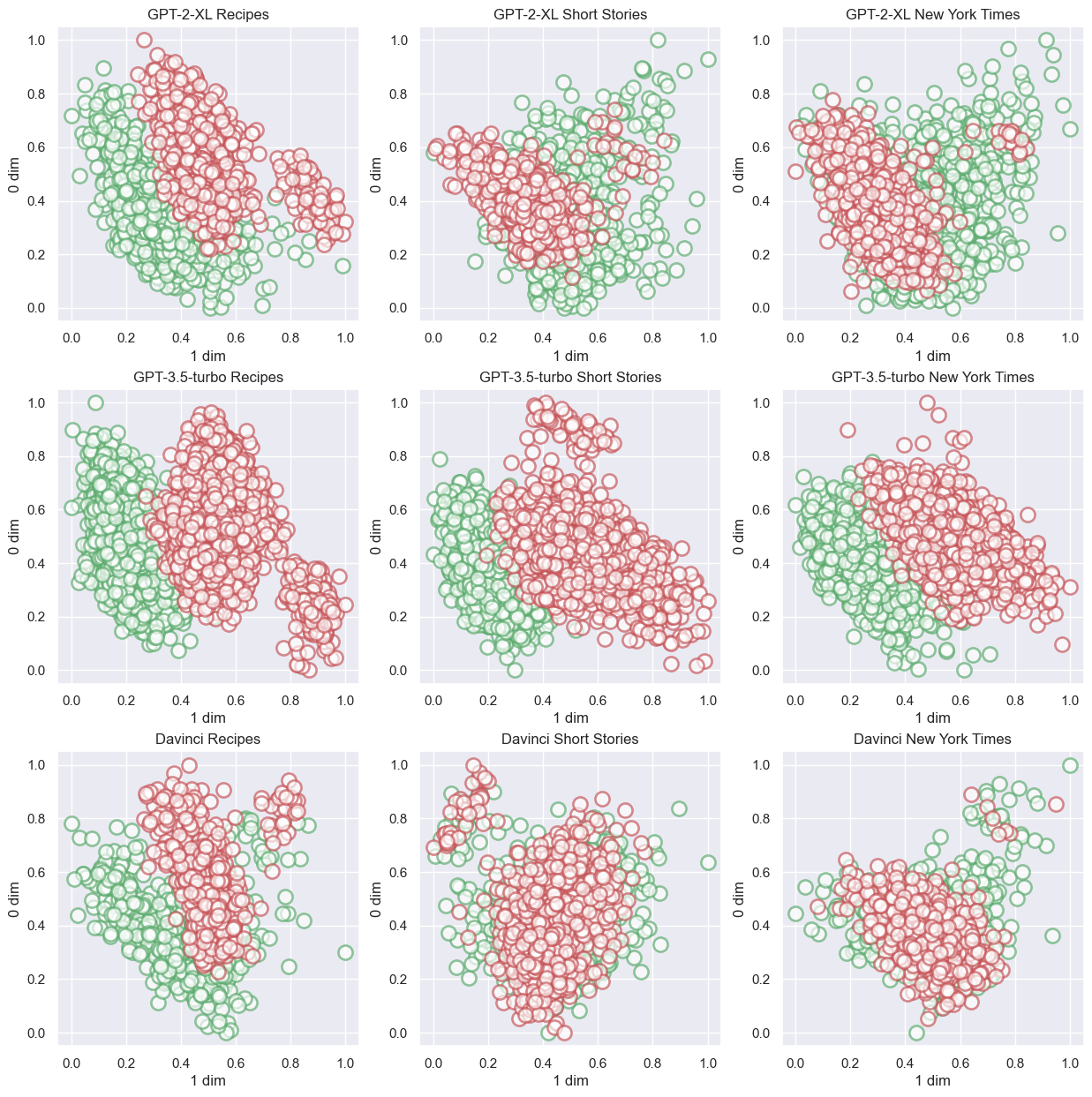}
    \caption{Distribution of pretrained (but not fine-tuned) RoBERTa [CLS] embeddings of real and fake parts of text samples from the original RoFT and \emph{RoFT-chatgpt} datasets. The dimension is reduced to 2D via principal component analysis~\citep{doi:10.1080/14786440109462720}.}
    \label{fig:embd_visualiz}

    \centering
    \includegraphics[width=\textwidth]{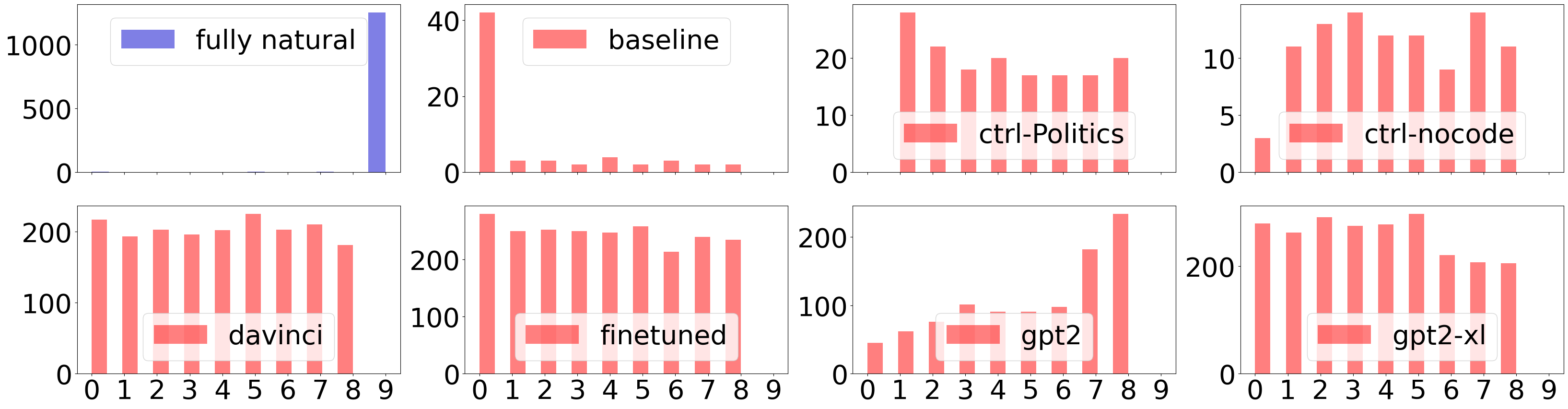}
    \caption{Label distributions for the original RoFT dataset by model}
    \label{fig:roft-labels-by-model}
    
    \centering
    \includegraphics[width=\textwidth]{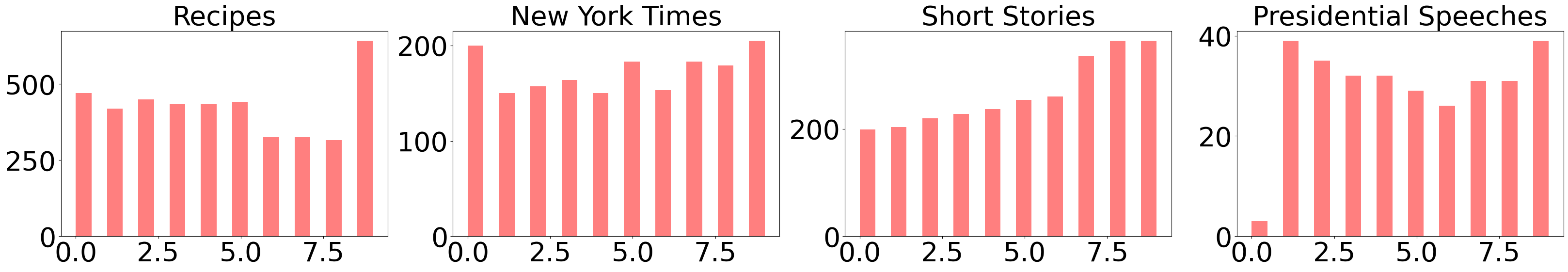}
    \caption{Label distributions for the original RoFT dataset by topic}
    \label{fig:roft-labels-by-topic}
\end{figure*}

\def\alittlespace{.6cm}

\begin{figure*}[!t]
    \centering
    \includegraphics[width=\textwidth]{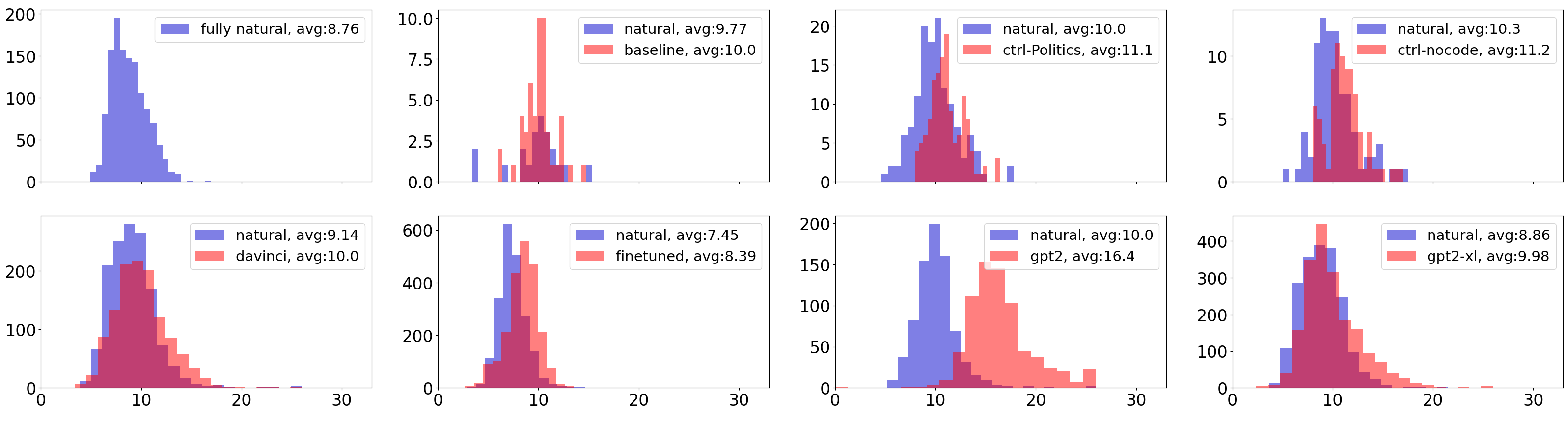}
    \caption{PHD distributions for the real and fake parts of the RoFT dataset, by generator models}
    \label{fig:hist_part_models}
    
    \centering
    \includegraphics[width=\textwidth]{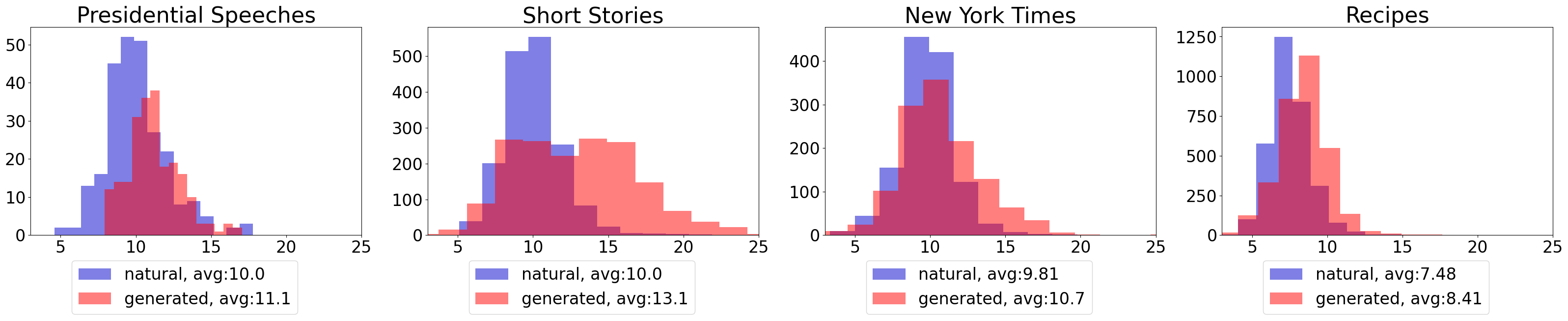}
    \caption{PHD distributions for the real and fake parts of the RoFT dataset, by topics}
    \label{fig:hist_part_topics}
    \vspace{\alittlespace}

    \centering
    \includegraphics[width=\textwidth]{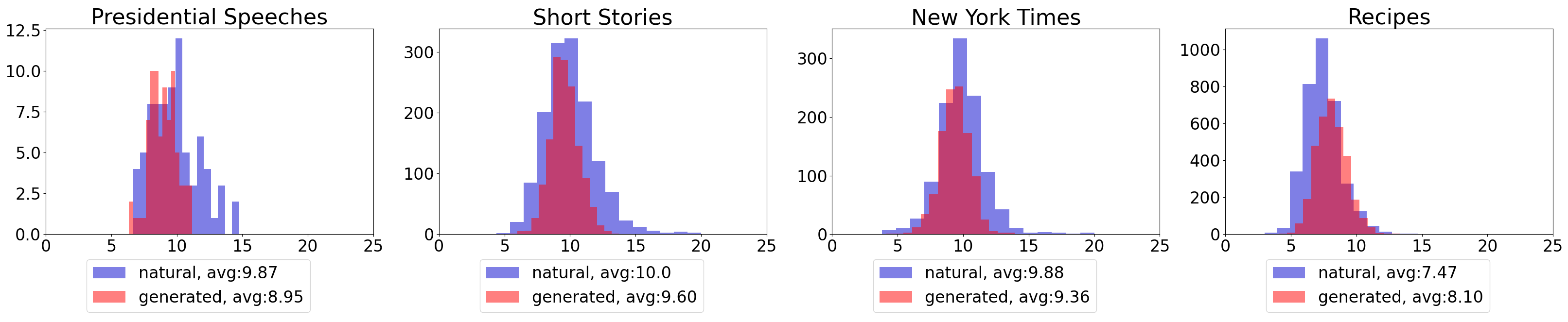}
    \caption{PHD distributions for the real and fake parts of the \emph{RoFT-chatgpt} dataset text, by topics}
    \label{fig:hist_part_topics_chat}
    \vspace{\alittlespace}
\end{figure*}

\begin{figure*}[!t]
    \centering
    \includegraphics[width=\textwidth]{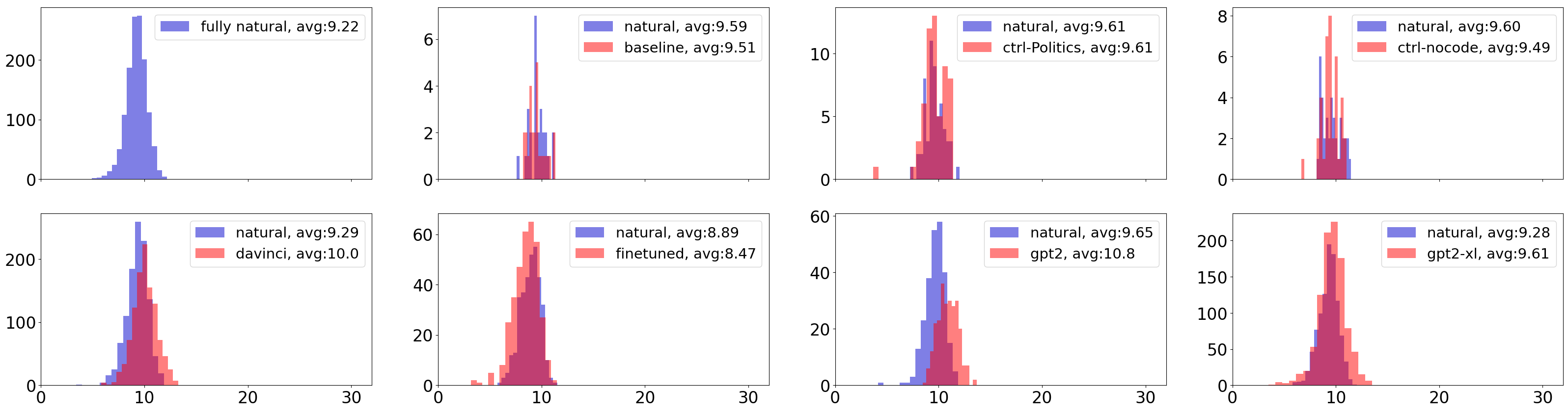}
    \caption{TLE dimension distributions for the sentences in the RoFT dataset by generator models}
    \label{fig:hist_sent_models}
    \vspace{\alittlespace}

    \centering
    \includegraphics[width=\textwidth]{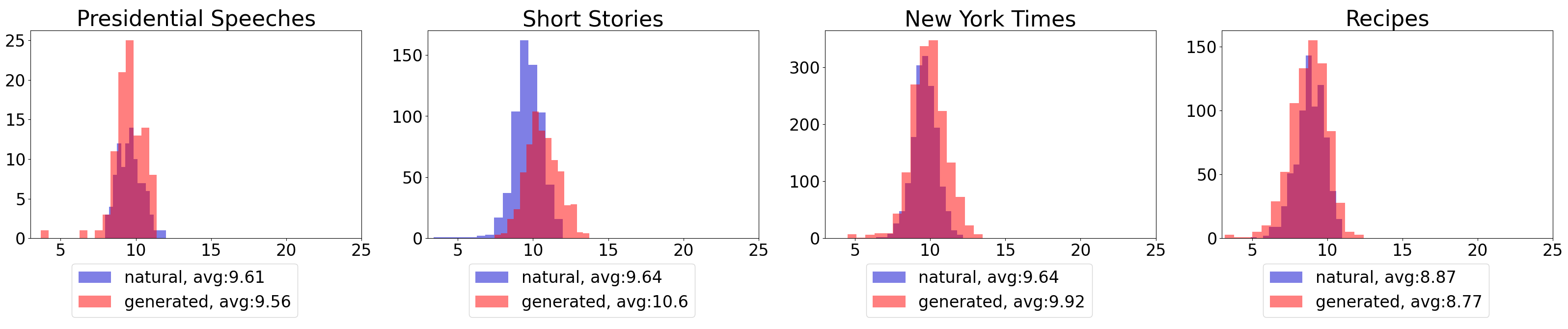}
    \caption{TLE dimension distributions for the sentences in the RoFT dataset by topics}
    \label{fig:hist_sent_topics}
    \vspace{\alittlespace}

    \centering
    \includegraphics[width=\textwidth]{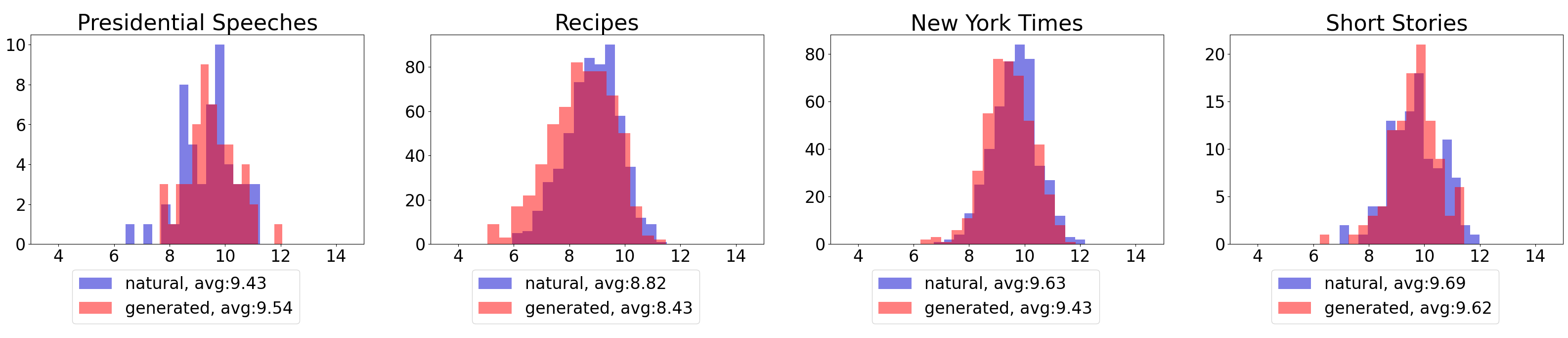}
    \caption{TLE dimension distributions for the sentences in the \emph{RoFT-chatgpt} dataset by topics}
    \label{fig:hist_sent_topics_chat}
\end{figure*}

In this section we provide additional statistics and visualizations for the distributions of various features in the data. In particular, we note that on most diagrams, real texts have smaller PHD than fake texts, which is a very different result from the statistics presented by \citet{tulchinskii2023intrinsic}, who noted that the PHD of real texts is larger than that of fake texts. We hypothesize that it can be due either to very short lengths of texts in our work compared to the texts considered by \citet{tulchinskii2023intrinsic} or due to differences in the sampling strategy used by \citet{dugan-etal-2020-roft} and \citet{tulchinskii2023intrinsic} when generating texts. Another observation is that the TLE dimension is very different for all generator models in the original RoFT dataset. This may be the reason for the bad generalization performance of intrinsic dimension-based algorithms across domains. For \emph{RoFT-chatgpt} PHD and TLE, real and fake texts are close to each other.

\begin{figure*}[!t]
    \centering
    \includegraphics[width=\textwidth]{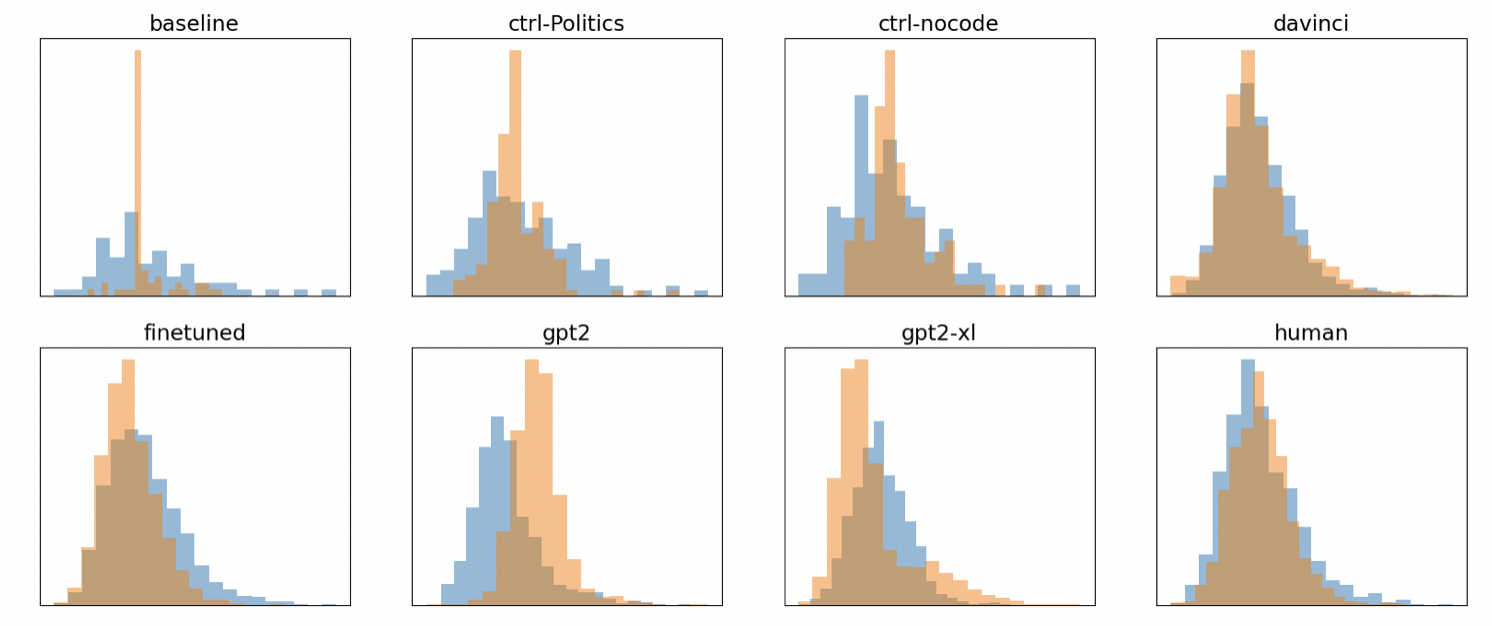}
    \caption{Perplexities of the last sentence in RoFT by model obtained from GPT-2; \emph{blue}~--- distribution on the in-domain set, i.e., the entire dataset except the speficied generator; \emph{orange}~--- on the out-of-domain set, i.e., data from the specified generator.}
    \label{fig:roft-dist-9th-by-model}
\end{figure*}

\begin{figure*}[!t]
    \centering
    \includegraphics[width=1\textwidth]{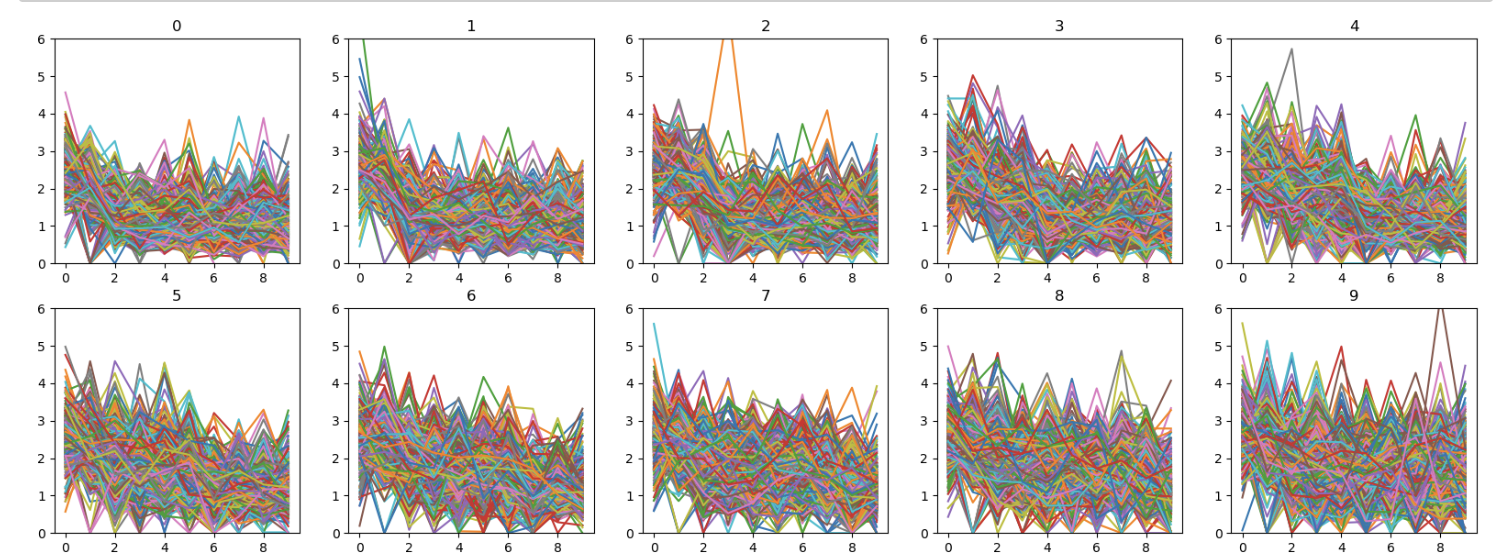}
    \caption{Sentence perplexities in the \emph{RoFT-chatgpt} dataset by label obtained from Phi-1.5 model. X axis: sentence index in the text, Y axis: sentence perplexity.}
    \label{fig:perp_by_sent_by_label}

    \centering
    \setlength{\tabcolsep}{2pt}
    \begin{tabular}{cc}
    \includegraphics[width=0.5\textwidth]{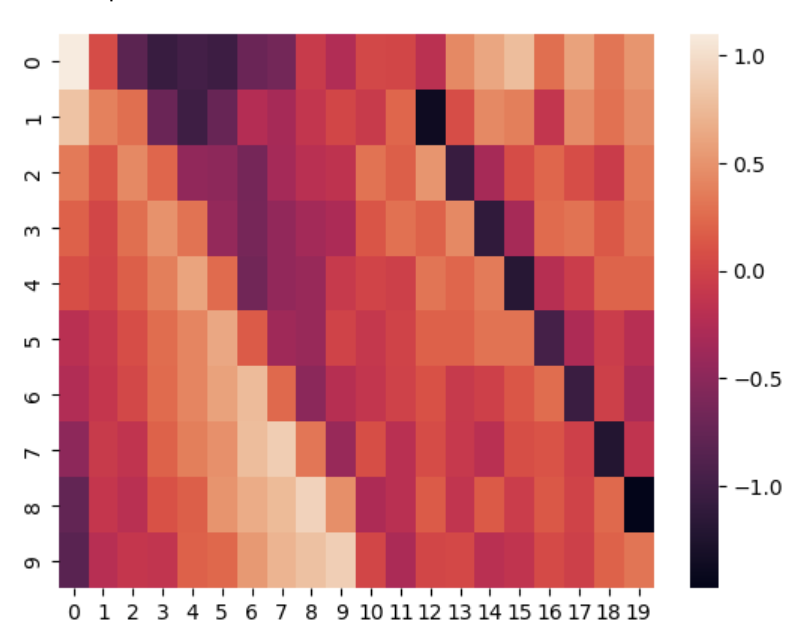} &
    \includegraphics[width=0.5\textwidth]{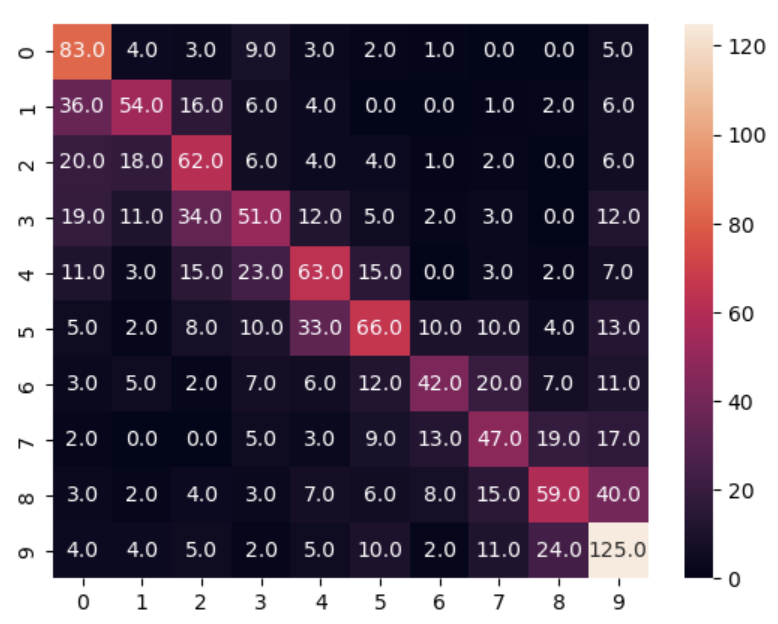} \\
    (a) & (b)
    \end{tabular}

    \caption{Analysis of the logistic regression trained on sentence perplexities on the \emph{RoFT-chatgpt} dataset (\emph{Perplexity + LogRegr} in the tables): (a) heatmap of the coefficients; (b) confusion matrix for test set predictions of logistic regression trained on sentence perplexities on the \emph{RoFT-chatgpt} dataset.}
    \label{fig:heatmaps}
\end{figure*}

We show dataset statistics in the following figures:
\begin{itemize}
    \item Figures~\ref{fig:roft-by-model}, \ref{fig:roft-by-topic}, and~\ref{fig:roft-chatgpt-by-topic} show the lengths of texts in tokens produced by the standard RoBERTa tokenizer (the figures have a cutoff of 100 for readability but the datasets do contain a few longer sentences);
    \item Figure~\ref{fig:embd_visualiz} shows the distribution of pretrained (but not fine-tuned) RoBERTa [CLS] embeddings for real and fake parts of text samples from the original RoFT and \emph{RoFT-chatgpt} datasets;
    \item Figure~\ref{fig:roft-labels-by-model} shows the distribution of labels in the original RoFT dataset by generator;
    \item Figure~\ref{fig:roft-labels-by-topic} shows the distribution of labels in the original RoFT dataset by topic; this distribution is identical to the corresponding distribution for the \emph{RoFT-chatgpt} dataset;
    \item Figure~\ref{fig:hist_part_models} shows the distribution of PH dimensions of real and fake parts of the text by generator;
    \item Figures~\ref{fig:hist_part_topics} and~\ref{fig:hist_part_topics_chat} show the distributions of PH dimensions by topic for the original RoFT and \emph{RoFT-chatgpt} respectively;
    \item Figure~\ref{fig:hist_sent_models} shows the distribution of TLE dimensions of different sentences by generator;
    \item Figures~\ref{fig:hist_sent_topics} and~\ref{fig:hist_sent_topics_chat} show the the distributions of TLE dimensions by topic for original RoFT and \emph{RoFT-chatgpt} respectively.
\end{itemize}

\section{Detailed experimental results}\label{sec:approaches_appendix}

In this section, we provide additional statistics and visualizations regarding our experimental results. Figure~\ref{fig:perp_by_sent_by_label} visualizes the changes in perplexities for sentences from the texts in \emph{RoFT-chatgpt} by their labels. We make the following observations.

First, perplexities of the first couple of sentences across all texts are quite high, and the average perplexity of sentences decreases by the end of the text. This is probably due to the fact that for the words of the first sentences the length of the text prefix is not enough for a stable calculation of perplexity. One solution to mitigate this effect and hence make perplexity-based classifiers more stable might be to generate new prefixes for the text using some generative model (e.g. \emph{gpt-3.5}) and calculate perplexities of original text words using this generated prefix. We leave this idea for further research.
Figure \ref{fig:heatmaps} visualizes the coefficients of a logistic regression model trained on sentence perplexities from the \emph{RoFT-chatgpt} dataset (\emph{Perplexity + LogRegr} rows in the tables). We can see a distinct pattern in this figure. For the label \emph{$k$}, which means that the first fake sentence in the text is the (\emph{$k+1$})st, the highest value of the coefficient are \emph{$k$}th and \emph{$k_{10}$}th ones, and the lowest ones are often (\emph{$k+2$})nd and (\emph{$k+12$})th. This could mean that the model is ``searching'' for a sudden drop of mean and variance of perplexity at a point where the fake part is starting. This fits together well with the idea that language model (Phi-1.5) trained on data from another model (GPT-3.5 version) see text generated by this model as a more ``natural'' one than real human-produced text. Therefore, perplexity often drops at the point where fake text begins, and logistic regression can pick up this effect and use it as a decision rule.

Finally, Figure~\ref{fig:heatmaps}b visualizes the confusion matrix on the test set of a logistic regression trained on the RoFT-ChatGPT dataset. We see that most of the errors concentrate around true labels, indicating that often model almost correctly finds a boundary, having a shift by +1/-1 label only.

\section{Examples of correctly classified and misclassified texts}\label{sec:misclassified_examples}

\begin{figure*}[!t]\centering
\includegraphics[width=0.7\linewidth]{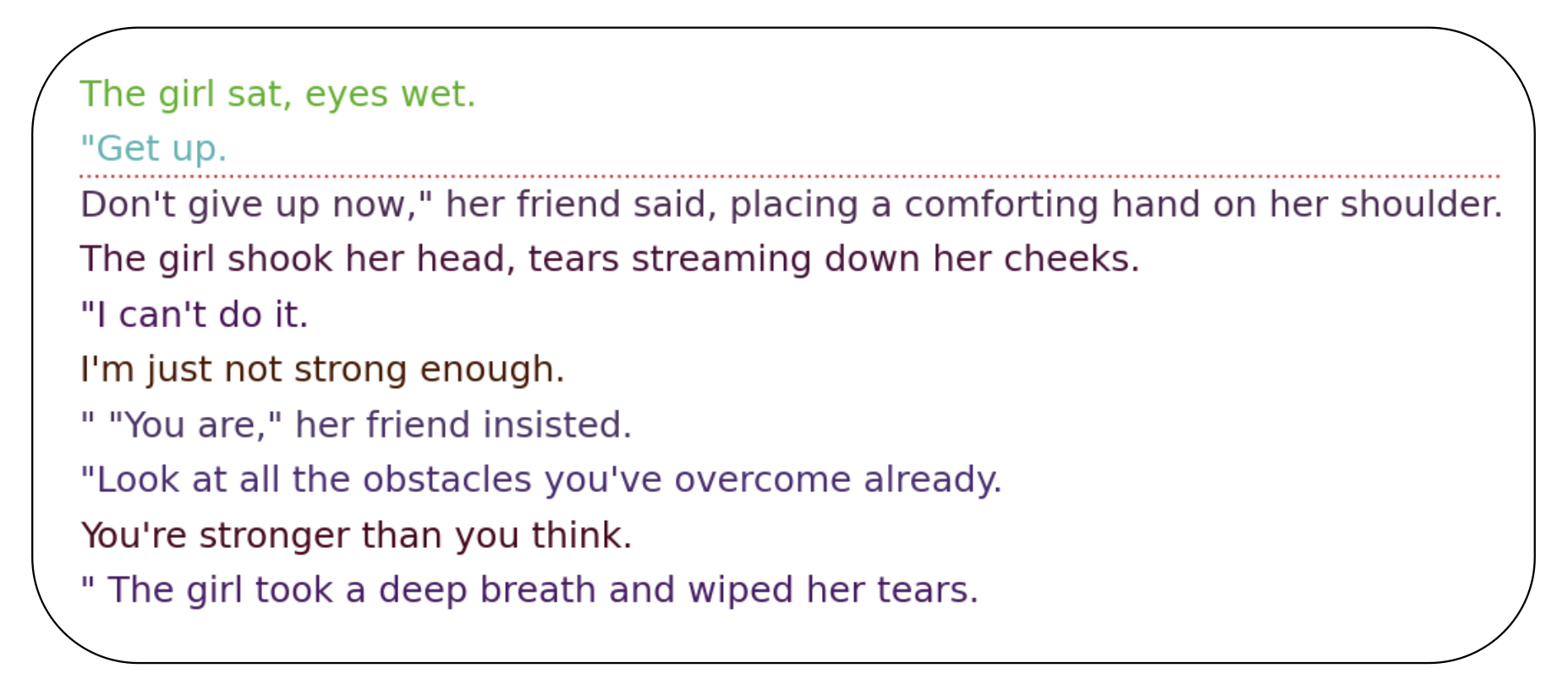}
\caption{Correctly classified sample input from the ROFT-chatgpt dataset colored according to the perplexities of Phi-1.5 on each sentence; ``greener'' text corresponds to higher mean perplexity, ``blue-er'', to higher standard deviation of the perplexity, red text shows where both mean and std are low. The prompt is above the dotted line, the rest is generated.}%
\label{fig:correct_example}
\end{figure*}

\begin{figure*}[!t]\centering
\includegraphics[width=\linewidth]{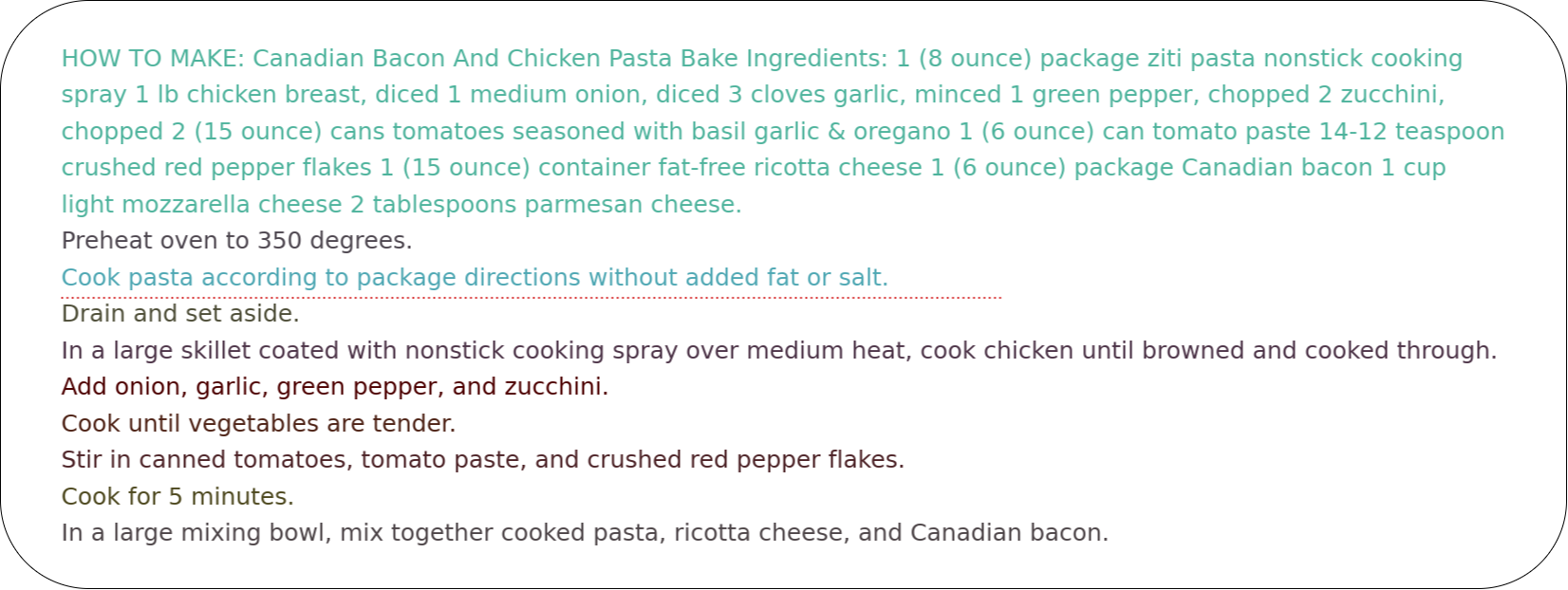}
\includegraphics[width=\linewidth]{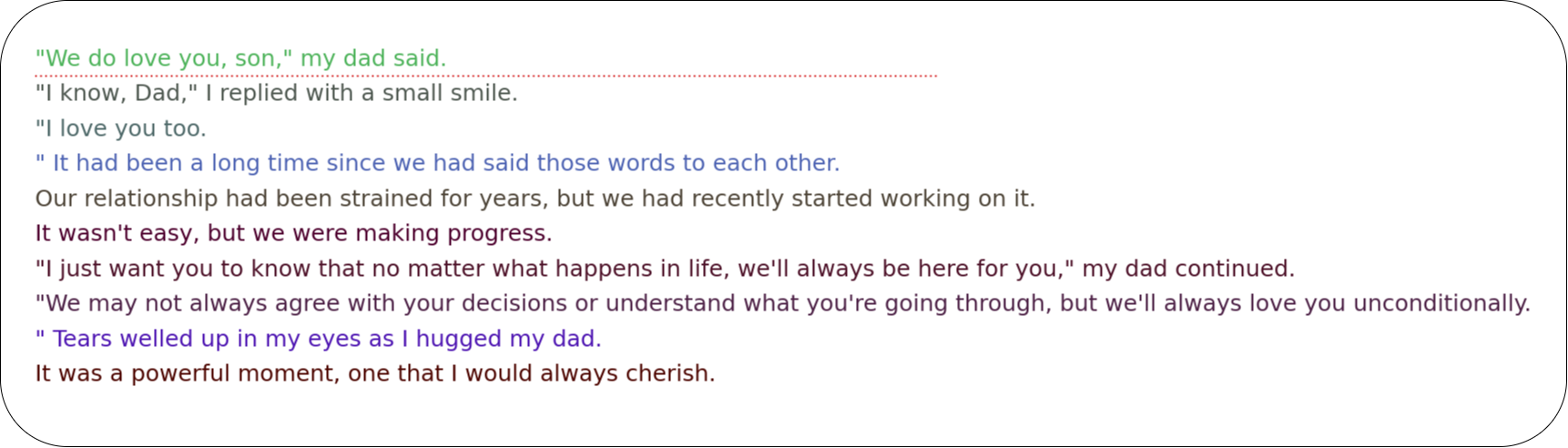}
\caption{Misclassified inputs from the ROFT-chatgpt dataset colored according to the perplexities of Phi-1.5 on each sentence. At the first picture, generated text starts from the fourth sentence, but perplexity-based classifier predicts that it starts from the fifth sentence; at the second picture, generated text starts from the second sentence, but perplexity-based classifier predicts that it starts from the fifth sentence.}%
\label{fig:incorrect_example}
\end{figure*}

We provided more examples of correctly and incorrectly classified texts at the Figures ~\ref{fig:correct_example} and \ref{fig:incorrect_example}.

\end{document}